\documentclass[journal]{IEEEtran}
\usepackage{amsmath,amsfonts}
\usepackage{algorithmic}
\usepackage{algorithm}
\usepackage{array}
\usepackage[caption=false,font=normalsize,labelfont=sf,textfont=sf]{subfig}
\usepackage{textcomp}
\usepackage{stfloats}
\usepackage{url}
\usepackage{verbatim}
\usepackage{graphicx}
\usepackage{cite}
\usepackage{booktabs}
\usepackage{multirow}
\usepackage{rotating}
\usepackage{makecell}
\usepackage{ifpdf}
\usepackage{color}
\usepackage{wasysym}
\usepackage{tabularx}

\begin{document}

\title{Brain-inspired AI for Edge Intelligence: a systematic review}
\author{Yingchao~Cheng, Meijia~Wang, Zhifeng~Hao,~\IEEEmembership{Senior Member,~IEEE}, Rajkumar Buyya,~\IEEEmembership{Fellow,~IEEE}
\thanks{This work was supported in part by the National Key R\&D Program of China (2025YFC3410000) and the Guangdong Basic and Applied Basic Research Foundation (2023B1515120020). (Corresponding author: Yingchao Cheng.)}
\thanks{Y. Cheng is with the Guangdong Laboratory of Artificial Intelligence and Digital Economy (SZ), Shenzhen, 518107, China. (e-mail: chengyingchao.dr@gmail.com).}
\thanks{M. Wang is with the Cloud Computing and Distributed Systems (CLOUDS) Lab, School of Computing and Information Systems, University of Melbourne, Melbourne, VIC 3010, Australia, and also with the Guangdong Laboratory of Artificial Intelligence and Digital Economy (SZ), Shenzhen, 518107, China. (e-mail: meijia.wang@student.unimelb.edu.au).}
\thanks{Z. Hao is with the College of Mathematics and Computer Science, Shantou University, Shantou, 515063, China. (e-mail: haozhifeng@stu.edu.cn)}
\thanks{R. Buyya is with the Cloud Computing and Distributed Systems (CLOUDS) Lab, School of Computing and Information Systems, University of Melbourne, Melbourne, VIC 3010, Australia (e-mail: rbuyya@unimelb.edu.au).}
}



\maketitle

\begin{abstract}
While Spiking Neural Networks (SNNs) promise to circumvent the severe Size, Weight, and Power (SWaP) constraints of edge intelligence, the field currently faces a ``Deployment Paradox'' where theoretical energy gains are frequently negated by the inefficiencies of mapping asynchronous, event-driven dynamics onto traditional von Neumann substrates. Transcending the reductionism of algorithm-only reviews, this survey adopts a rigorous \textit{system-level hardware-software co-design perspective} to examine the 2020--2025 trajectory, specifically targeting the ``last mile'' technologies—from quantization methodologies to hybrid architectures—that translate biological plausibility into silicon reality. We critically dissect the interplay between training complexity (the dichotomy of direct learning vs. conversion), the ``memory wall'' bottlenecking stateful neuronal updates, and the critical software gap in neuromorphic compilation toolchains. Finally, we envision a roadmap to reconcile the fundamental ``Sync-Async Mismatch,'' proposing the development of a standardized \textit{Neuromorphic OS} as the foundational layer for realizing a ubiquitous, energy-autonomous \textit{Green Cognitive Substrate}.
\end{abstract}


\section{Introduction}

\IEEEPARstart{T}{he} exponential proliferation of Internet of Things (IoT) devices has catalyzed a seismic paradigm shift, migrating Artificial Intelligence (AI) from hyperscale cloud infrastructures to the resource-constrained network edge \cite{shi2016edge, zhou2019edge, buyya2018manifesto}. This migration is driven not merely by preference but by necessity; we are currently witnessing a global ``data deluge,'' where the sheer velocity and volume of sensory data generation are outpacing available transmission bandwidth. With global data creation projected to reach zettabyte scales, the traditional cloud-centric model faces critical bottlenecks: \textbf{bandwidth saturation} and unpredictable transmission latency. Relying on centralized servers to process raw, high-fidelity streams (e.g., 4K video surveillance or autonomous driving LiDAR) is no longer sustainable, necessitating a push towards ``Edge Intelligence'' where data is processed in situ \cite{rydning2018digitization, chen2019deep, satyanarayanan2017emergence, botta2016integration}.

However, this transition faces a fundamental physical impasse: the deployment of traditional Deep Neural Networks (DNNs) on battery-operated edge devices is increasingly constrained by the \textbf{von Neumann bottleneck}. The inherent requirement of DNNs for continuous, dense matrix multiplications necessitates massive and frequent data shuttling between physically separated memory and processing units. This creates a ``memory wall'' that induces latency and power consumption metrics often prohibitive for real-time, mission-critical applications \cite{Ref_VonNeumann_EdgeAI1, Ref_VonNeumann_EdgeAI2, Ref_VonNeumann_EdgeAI3}.

\begin{figure*}[t]
    \centering
    \includegraphics[width=0.9\textwidth]{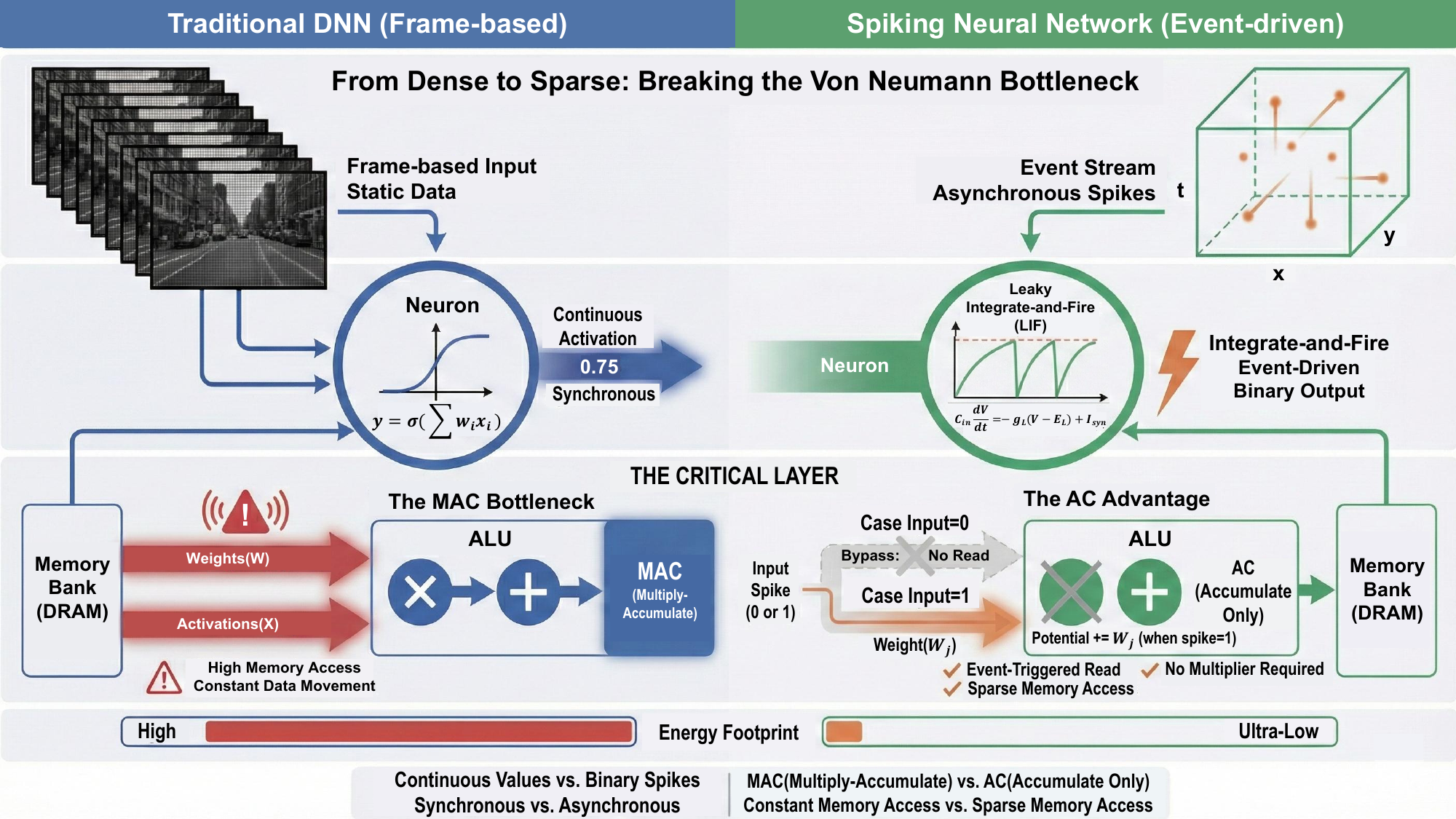}
    \vspace{-2mm}
    \caption{\textbf{Overview of the paradigm shift from Traditional DNNs to Spiking Neural Networks (SNNs).} 
    \textbf{(Top) Data Representation:} DNNs process redundant frame-based image sequences, whereas SNNs process sparse, asynchronous event streams represented in a spatiotemporal volume ($x, y, t$). 
    \textbf{(Middle) Neuron Dynamics:} Unlike static activation functions (e.g., Sigmoid) in DNNs, SNNs employ biologically plausible Leaky Integrate-and-Fire (LIF) neurons that integrate temporal information governed by differential equations.
    \textbf{(Bottom) Energy Efficiency:} The shift from dense MAC operations to event-driven Accumulate (AC) logic results in an ultra-low energy footprint, making SNNs highly suitable for edge intelligence.}
    \label{fig:paradigm}
\end{figure*}

In response to these hardware limitations, \textbf{Spiking Neural Networks (SNNs)} have transcended their origins as biological curiosities to become a pragmatic computational imperative for the next generation of Edge Intelligence. Diverging from the frame-based, synchronous processing of conventional DNNs, SNNs operate on an asynchronous, event-driven paradigm that mirrors the efficiency of biological neural substrates. By encoding information into discrete binary spikes, SNNs leverage extreme \textbf{spatio-temporal sparsity} to decouple computation from the clock cycle—effectively achieving a state of ``computational parsimony''. This mechanism fundamentally transforms the dominant workload from power-hungry Multiply-Accumulate (MAC) operations to significantly more energy-efficient Accumulate (AC) operations \cite{Ref_SNN_Sparsity_Efficiency1, Ref_SNN_Sparsity_Efficiency2, Ref_SNN_Sparsity_Efficiency3, Ref_SNN_Sparsity_Efficiency4}, as illustrated in Fig.~\ref{fig:paradigm}. \textit{Consequently, SNNs represent not just an algorithmic optimization, but a structural alignment between AI models and the sparse, event-driven nature of real-world sensory data.}

Nevertheless, bridging the chasm between theoretical algorithmic efficiency and physical hardware deployment remains a critical challenge. The field is currently encumbered by a \textbf{``Deployment Paradox''}: while SNN algorithms demand event-driven substrates, the dominant commercial off-the-shelf (COTS) edge hardware—such as NVIDIA Jetson series or Edge TPUs—remains architecturally optimized for dense, synchronous matrix operations. This inefficiency stems from a fundamental mismatch in \textbf{Single Instruction, Multiple Data (SIMD)} parallelism. GPUs achieve peak throughput via lockstep execution of dense threads and coalesced memory access; however, the stochastic, irregular firing patterns of SNNs introduce severe control-flow divergence (breaking SIMD lockstep) and non-contiguous memory requests. This \textbf{Memory Access Granularity} mismatch implies that fetching a single weight for a sparse spike often triggers the transfer of an entire cache line, resulting in low effective bandwidth utilization. Consequently, mapping sparse SNNs onto these general-purpose accelerators forces the use of inefficient simulation layers that frequently negate the inherent energy benefits of spiking models \cite{LiuYanzhen2024, Ref_COTS_SNN_Inefficiency2, Ref_COTS_SNN_Inefficiency3}.

Conversely, native neuromorphic processors capable of unlocking the full potential of SNNs, such as Intel Loihi 2 or BrainChip Akida, operate on the requisite asynchronous paradigms but are largely restricted to research prototypes or niche vertical markets. This dichotomy creates a significant \textbf{deployment gap}, where advanced algorithms lack suitable, ubiquitous hardware hosts \cite{ZhangGuanlei2025, Ref_Neuromorphic_Hardware1, Ref_Neuromorphic_Hardware2, Ref_Neuromorphic_Hardware3}. \textit{Ultimately, the widespread adoption of edge SNNs is currently stalled not by algorithmic incapacity, but by the inertia of a hardware ecosystem optimized for the previous generation of dense deep learning models.}

While prior surveys have attempted to chart this territory, the existing literature has predominantly bifurcated the landscape, treating SNN algorithms and hardware implementations as distinct, isolated domains. One category of reviews focuses heavily on algorithmic innovations—such as surrogate gradient learning or plasticity rules—while abstracting away the physical constraints of deployment platforms \cite{schuman2022opportunities, Eshraghian2023Survey}. Conversely, hardware-centric surveys often delve into emerging devices (e.g., memristors) or circuit designs without adequately addressing the scalability of the algorithms required to run on them \cite{ZhouChenlin2024, Ref_Existing_General_Reviews, ShenShuaijie2024}. This separation obscures the practical intricacies of the deployment gap, leaving a void in understanding how algorithmic sparsity interacts with hardware primitives. To bridge this divide, the present article adopts a holistic \textbf{hardware-software co-design perspective}. \textit{This approach posits that the efficacy of Edge Intelligence is not determined solely by synaptic precision, but by the seamless orchestration of algorithmic sparsity and hardware architecture.} 

Accordingly, this work articulates three primary contributions:

\begin{itemize}
    \item \textbf{Critical Synthesis of Training Paradigms (2020--2025):} We provide a comparative analysis of recent breakthroughs in direct training mechanisms (e.g., Surrogate Gradients) versus ANN-to-SNN conversion techniques. The discussion evaluates these methodologies not in isolation, but through the lens of edge suitability—balancing training latency against inference energy efficiency \cite{NEURIPS2022_82846e19, bu2023optimal, NEURIPS2024_7c9341ad, zhu2023spikegpt, zhou2024spikformer}.

    \item \textbf{Deployment-Centric System Analysis:} Distinguished from generic algorithmic overviews, this survey addresses the ``last mile'' challenges of physical realization. We scrutinize the often-overlooked bottlenecks of post-training quantization, heterogeneous mapping toolchains, and inter-chip communication overheads that govern real-world performance \cite{snyder2023neuromorphic, pedersen2024neuromorphic, rasmussen2019nengodl, carpegna2024spiker}.

    \item \textbf{Prospective Roadmap for Connected Intelligence:} Extending beyond standard computer vision benchmarks, the scope is broadened to emerging frontiers. We delineate the role of SNNs in distributed edge learning frameworks and their potential integration with next-generation 6G networks, identifying the trajectory towards ubiquitous, ultra-low-power cognitive radio systems \cite{km2025spiking, huang2024attention, wang2024snn}.
\end{itemize}

The structural organization of this survey is distinctively predicated on the principle of \textbf{Hardware-Software Co-design}. Departing from the dichotomy observed in extant literature—where algorithmic sparsity and neuromorphic substrates are often dissected in isolation—this work posits that bridging the ``last mile'' of edge deployment necessitates a unified taxonomy. Under this framework, algorithmic parameters (e.g., sparsity levels, neuron models) are treated as inextricably linked to hardware constraints (e.g., fan-in limits, memory hierarchy), thereby establishing a cohesive roadmap for ubiquitous edge intelligence \cite{svoboda2025spiking, shao2023efficient, aliyev2024sparsity}. 

\begin{figure*}[t]
    \centering
    \includegraphics[width=0.7\linewidth]{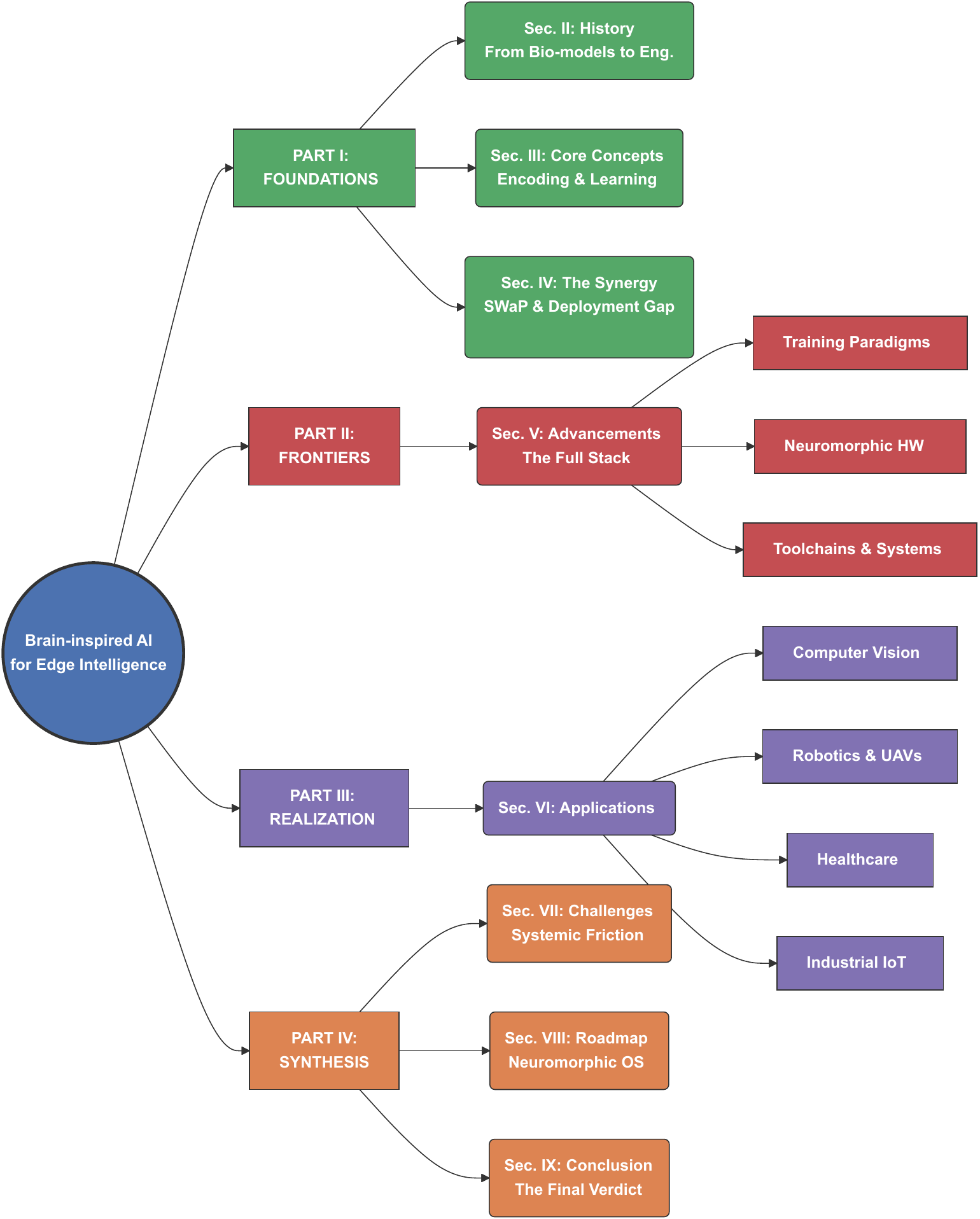}
    \vspace{-2mm} 
    \caption{\textbf{Hierarchical Organization of the Survey.} The article is structured into four main pillars: \textbf{Part I} establishes the theoretical foundations and defines the ``Deployment Paradox''; \textbf{Part II} explores the technical frontiers across algorithms, hardware, and toolchains (2020--2025); \textbf{Part III} examines real-world realization in key edge applications; and \textbf{Part IV} synthesizes challenges to propose a roadmap towards a unified Neuromorphic OS.}
    \label{fig:structure}
\end{figure*}

To systematically unravel these interdependencies, the remainder of this article is structured as follows (visually summarized in Fig.~\ref{fig:structure}):

\begin{itemize}
    \item \textbf{Foundations and Motivation (Sections \ref{sec:history}--\ref{sec:edge_suitability}):} The narrative begins by tracing the evolutionary trajectory of SNNs from biophysical roots to modern computational primitives (Section \ref{sec:history}). This is followed by a rigorous definition of core concepts, including encoding schemes and a comparative analysis against DNNs (Section \ref{sec:concepts}). Subsequently, Section \ref{sec:edge_suitability} elucidates the intrinsic alignment between SNN dynamics and the energy-latency constraints of edge environments.

    \item \textbf{Technological Frontier (Section \ref{sec:advancements}):} Representing the core technical synthesis (2020--2025), this section comprehensively evaluates the state-of-the-art across the full stack: from novel training paradigms and hardware-aware optimizations to the mapping toolchains required to bridge the software-hardware gap.

    \item \textbf{Realization and Reflection (Sections \ref{sec:applications}--\ref{sec:conclusion}):} The theoretical discussion transitions into practice in Section \ref{sec:applications}, exploring deployment in computer vision, robotics, and IoT. Section \ref{sec:challenges} then adopts a critical lens to examine persistent bottlenecks in scalability and training. Finally, Section \ref{sec:future} projects the field's trajectory towards 6G integration and ethical AI, before concluding the survey in Section \ref{sec:conclusion}.
\end{itemize}

\textit{Ultimately, this structure is designed to guide the reader from microscopic neuron dynamics to macroscopic system-level deployment, highlighting that the future of edge AI lies in the optimization of the interface between these two scales.}

\section{Background and Foundations: The SNN Paradigm}
\label{sec:history}

The theoretical trajectory of Spiking Neural Networks (SNNs) is defined by a central dichotomy: the pursuit of biological isomorphism versus the necessity of engineering tractability. Unlike the continuous activation landscapes of second-generation Artificial Neural Networks (ANNs), SNNs operate on the premise of discrete, event-driven information processing. This section delineates the evolution of this paradigm, tracing the shift from rigorous biophysical modeling to the modern era of algorithmic differentiation and hardware co-design.

\subsection{Modeling Dynamics: From Ion Channels to Abstraction}
The classification of SNNs as the \textit{third generation} of neural networks is predicated on their ability to encode information in the precise timing of spikes, theoretically offering superior computational power over rate-based counterparts \cite{Maass1997, maass2004computational, izhikevich2004which}. However, realizing this potential required a fundamental compromise in modeling granularity.

The roots of neuronal modeling are established in the rigorous description of action potential generation via ion channel conductance, typified by the Hodgkin-Huxley formalism \cite{hodgkin1952quantitative, markram2015reconstruction, poirazi2020illuminating}. While biophysically accurate, such high-dimensional differential equations proved prohibitively expensive for network-level simulations. This computational bottleneck necessitated a strategic pivot toward phenomenological abstraction. The Leaky Integrate-and-Fire (LIF) model emerged as the standard for efficient simulation, reducing dynamics to a linear membrane integration process reset by a hard threshold \cite{Gerstner2002, teeter2018generalized, skatchkovsky2020federated, lobo2020spiking}. 

To bridge the gap between the simplistic LIF and the complex Hodgkin-Huxley models, hybrid approaches were developed. Notably, models that capture diverse firing patterns (e.g., bursting, chattering) through bifurcations in reduced-dimensional phase systems offered a critical trade-off, enabling rich dynamics with the computational cost of simple maps \cite{Izhikevich2003, brette2005adaptive, naud2008firing, jolivet2008benchmark}.

Parallel to these algorithmic abstractions, the hardware substrate underwent a similar philosophical shift. The concept of ``neuromorphic'' engineering was originally conceived to exploit the physics of subthreshold analog VLSI to directly emulate neurobiological ion flow \cite{mead1989analog, mead1990neuromorphic, indiveri2011neuromorphic}. Over time, this evolved into digital and mixed-signal implementations, prioritizing the energy efficiency of sparse, event-driven communication over strict analog emulation.

\vspace{0.5em}
\noindent \textit{The evolution of neuron models illustrates a clear trend toward functional minimalism. The field has largely converged on the realization that for neuromorphic computing, capturing the ``essential nonlinearity'' of the spike mechanism is more critical than replicating the precise ionic currents of the biological soma.}

\subsection{The Renaissance: Overcoming the Differentiability Barrier}
Despite a strong theoretical bedrock, the practical deployment of SNNs was historically impeded by the non-differentiable nature of the spike generation function, which rendered standard backpropagation ineffective. 

The initial era of SNN learning was dominated by biologically grounded, local learning rules, most notably Spike-Timing-Dependent Plasticity (STDP) and Hebbian variations \cite{markram1997regulation, song2000competitive, bi1998synaptic}. While these unsupervised mechanisms excelled at local feature extraction and pattern recognition in shallow architectures, they struggled to scale to the supervised complexity required for deep, hierarchical representations comparable to modern Convolutional Neural Networks (CNNs) \cite{neftci2019surrogate, richards2019deep, tavanaei2019deep,roy2019towards}.

The contemporary renaissance of SNNs—and their subsequent proliferation in edge computing—was catalyzed by the formalization of Surrogate Gradient (SG) methods. By approximating the non-differentiable Dirac delta function with a smooth auxiliary function during the backward pass, SG paradigms enabled the direct application of gradient descent to temporal dynamics, effectively linking the energy efficiency of SNNs with the training efficacy of Deep Learning \cite{Wu2018STBP, li2021differentiable, bellec2020solution,zenke2018superspike,shrestha2018slayer}. 

This algorithmic breakthrough coincided with the maturation of industrial-scale neuromorphic hardware \cite{merolla2014million, davies2018loihi, Pei2019Tianjic, furber2014spinnaker}. The validation that event-driven architectures could achieve orders-of-magnitude power reduction over von Neumann systems provided the physical imperative for deploying these deep spiking architectures in resource-constrained environments.

\vspace{0.5em}
\noindent \textit{The resurgence of SNNs is not merely a return to biological inspiration, but a pragmatic response to the ``memory wall'' and power constraints of modern computing. The shift from pure STDP to Gradient-based learning represents a compromise: accepting non-biological training methods to achieve biological levels of energy efficiency in inference.}

\section{Core Concepts of SNNs: An Engineering Perspective}
\label{sec:concepts}

SNNs represent a paradigm shift from continuous-valued activation to discrete, event-driven computation. Unlike conventional ANNs that rely on \textbf{frame-based, synchronous} processing, SNNs leverage the temporal dynamics of spike trains to encode information. This fundamental difference enables asynchronous processing and significant energy reductions, particularly in edge scenarios where power budgets are strictly constrained.

\subsection{Spiking Neuron Models: Fidelity vs. Cost}
The selection of a neuron model in neuromorphic engineering is a strategic trade-off between bio-fidelity and computational efficiency. Models range from biophysically accurate descriptions of ionic channels to simplified abstractions optimized for large-scale hardware implementation \cite{Yamazaki2022, izhikevich2004which, ZhouChenlin2024, svoboda2025spiking}.

\subsubsection{Leaky Integrate-and-Fire (LIF): The Simulation Standard}
The LIF model is the widely adopted abstraction for large-scale SNN simulations due to its linear subthreshold dynamics. It captures the essential ``integration'' and ``firing'' mechanism without modeling complex channel kinetics. Mathematically, it is described as:
\begin{equation}
\tau_m \frac{dV(t)}{dt} = -(V(t) - V_{\text{rest}}) + R_m I_{\text{syn}}(t),
\end{equation}
where $\tau_m = R_m C_m$ is the membrane time constant. While simplistic, the LIF model provides a convex optimization landscape for surrogate gradient learning, making it the \textit{de facto} standard for Deep SNNs \cite{karamimanesh2025fpga}.

\subsubsection{Izhikevich Model: The Computational Bridge}
The Izhikevich model serves as a computational bridge, capturing diverse firing patterns (e.g., bursting, chattering) typically associated with complex biophysical models, yet retaining the efficiency of coupled first-order differential equations:
\begin{equation}
\begin{aligned}
\frac{dv}{dt} &= 0.04v^2 + 5v + 140 - u + I, \\
\frac{du}{dt} &= a(bv - u).
\end{aligned}
\end{equation}
where $v$ is the membrane potential, $u$ is the recovery variable, and $I$ is the input current. By tuning parameters $a, b, c, d$, this model reproduces cortical dynamics with floating-point operations (FLOPs) comparable to simple integrators \cite{Iaboni2024}.

\subsubsection{Hodgkin–Huxley (HH): The Biophysical Gold Standard}
The HH model explicitly describes ionic conductance (Na$^+$, K$^+$). While it remains the gold standard for understanding neurophysiology, its high computational cost (requiring the solution of four non-linear differential equations) renders it impractical for edge intelligence applications, serving primarily as a validation benchmark \cite{roy2019towards, brette2007simulation}.

\vspace{0.5em}
\noindent \textit{The engineering objective in SNN design is not maximizing biological realism, but maximizing the ratio of information processing capability to energy consumption. Consequently, simple models like LIF dominate hardware implementations because they minimize the silicon area required per neuron, allowing for massive parallelism.}

\subsection{Encoding Taxonomies: Rate vs. Temporal}
Information encoding in SNNs dictates the system's latency and bandwidth efficiency. These schemes generally fall into two categories: rate-based and temporal coding \cite{thorpe2001spike, gautrais1998rate, mostafa2017supervised, auge2021survey}.

\subsubsection{Rate-Based Encoding}
Derived from classical neurophysiology, rate coding represents information via the mean firing frequency over a time window. While it offers high robustness against noise and stochasticity, it inherently incurs high latency (requiring time to average) and reduced energy efficiency due to redundant spiking.

\subsubsection{Temporal Encoding}
Temporal schemes exploit the precise timing of spikes to carry information, offering superior sparsity and bandwidth efficiency. 
\textbf{Time-to-First-Spike (TTFS)} encodes information inversely to the latency of the first spike, allowing for ultra-fast decision-making often termed ``latency coding'' \cite{Malcolm2023, svoboda2025spiking}.
Other methods include \textbf{Phase Coding}, where spikes are timed relative to a periodic background oscillation, and \textbf{Delta Modulation}, which is widely used in asynchronous sensors (e.g., DVS cameras) to encode only intensity changes, naturally filtering out static redundancy \cite{Liang2025, guo2021neural}.

\vspace{0.5em}
\noindent \textit{The transition from Rate to Temporal coding represents the critical leap from ``emulating biology'' to ``exploiting sparsity.'' Temporal codes maximize the information content per bit (spike), which is the theoretical foundation for SNNs' ultra-low power consumption.}

\subsection{Learning Dynamics: Local to Global}
Learning in SNNs is bifurcated into biologically plausible local rules and performance-oriented global optimization.

\subsubsection{Local Unsupervised Learning (STDP)}
Spike-Timing-Dependent Plasticity (STDP) modifies synaptic weights based on the causal latency $\Delta t = t_{\text{post}} - t_{\text{pre}}$. It enables self-organizing feature extraction without labeled data:
\begin{equation}
\Delta w = 
\begin{cases} 
A_+ e^{-\Delta t / \tau_+} & \text{if } \Delta t > 0 \text{ (LTP)} \\
-A_- e^{\Delta t / \tau_-} & \text{if } \Delta t < 0 \text{ (LTD)}
\end{cases}
\end{equation}
Variants such as R-STDP (Reward-modulated) and mSTDP (Mirrored) have been developed to introduce supervision signals into this local process \cite{ShenShuaijie2024, jo2010nanoscale, frenkel2020bottom}.

\subsubsection{Global Supervised Learning (Surrogate Gradients)}
To overcome the non-differentiability of binary spikes, modern SNNs utilize Surrogate Gradient (SG) methods. This approach smooths the Heaviside step function during backward propagation, allowing SNNs to be trained with the same efficacy as DNNs using backpropagation through time (BPTT) \cite{neftci2019surrogate, ZhouChenlin2024, gygax2025elucidating}.

\vspace{0.5em}
\noindent \textit{A convergence is emerging where ``local'' plasticity (STDP) is used for efficient pre-training or hardware adaptation, while ``global'' gradients (SG) fine-tune the network for task-specific accuracy. This hybrid approach balances the autonomy of biological systems with the precision of engineering control.}

\subsection{Architectural Comparison: SNNs vs. DNNs}
\begin{figure}[htb]
    \centering
    \includegraphics[width=1.0\columnwidth]{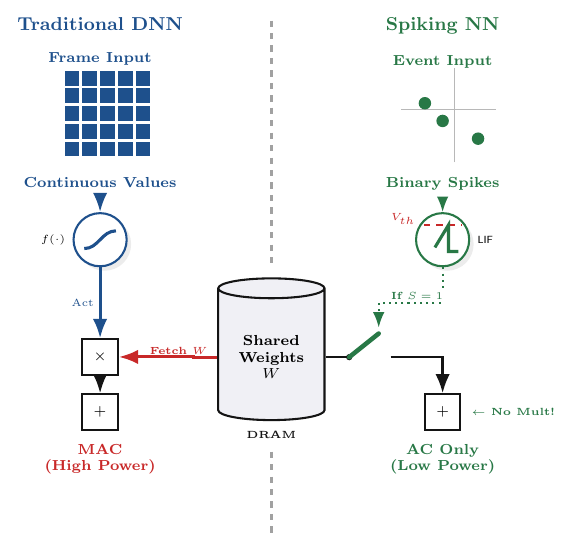}
    \caption{\textbf{Dataflow Contrast: Continuous vs. Event-Driven.} 
    \textbf{(Left) DNN Dataflow:} Relies on dense Matrix-Vector Multiplication (MAC), creating a continuous ``Memory Wall'' as weights are fetched unconditionally. 
    \textbf{(Right) SNN Dataflow:} Utilizes a conditional trigger mechanism where memory access and computation occur \textit{only} upon spike arrival. This sparsity-aware processing significantly reduces off-chip memory bandwidth requirements.}
    \label{fig:hardware_dataflow}
\end{figure}

Table \ref{tab:snn_dnn_comparison} and Fig.~\ref{fig:hardware_dataflow} synthesize the architectural distinctions. The decisive advantage of SNNs lies in replacing Multiply-and-Accumulate (MAC) operations with Accumulate (AC) operations. Since spikes are binary ($S(t) \in \{0, 1\}$), the synaptic weighting becomes a conditional addition: $V_{j}(t) = V_{j}(t-1) + \sum_{i} w_{ij} \cdot S_{i}(t)$. This bypasses the need for high-precision multipliers, reducing silicon footprint and dynamic power consumption \cite{HuoBingqiang2025, WeiWenjie2024}.

\begin{table*}[t]
\centering
\caption{Comparison of SNNs and DNNs} 
\label{tab:snn_dnn_comparison}

\renewcommand{\arraystretch}{1.3}

\begin{tabularx}{\linewidth}{@{} l >{\raggedright\arraybackslash}X >{\raggedright\arraybackslash}X @{}}
\toprule
\textbf{Feature} & \textbf{SNNs} & \textbf{DNNs} \\
\midrule
\textbf{Neuron Model} 
    & Biologically inspired (e.g., LIF, Izhikevich); discrete spikes 
    & Abstracted; continuous activation values (e.g., ReLU, sigmoid) \\ 
    \addlinespace[4pt] 

\textbf{Information Encoding} 
    & Temporal patterns of spikes (rate, latency, phase, population) 
    & Magnitude of activations \\
    \addlinespace[4pt]

\textbf{Computation} 
    & Event-driven, asynchronous, sparse; primarily Accumulate (AC) ops 
    & Synchronous, dense; primarily Multiply-Accumulate (MAC) ops \\
    \addlinespace[4pt]

\textbf{Energy Efficiency} 
    & Potentially very high, especially on neuromorphic hardware 
    & Generally lower, can be power-intensive \\
    \addlinespace[4pt]

\textbf{Temporal Processing} 
    & Intrinsic capability to process temporal data 
    & Often requires recurrent architectures (RNNs, LSTMs) \\
    \addlinespace[4pt]

\textbf{Biological Plausibility} 
    & Higher; mimics neural dynamics and learning rules like STDP 
    & Lower; learning (e.g., backpropagation) often not bio-plausible \\
    \addlinespace[4pt]

\textbf{Training Complexity} 
    & Challenging due to non-differentiable spikes; STDP, surrogate grads 
    & Well-established (backpropagation), but can be data-hungry \\
    \addlinespace[4pt]

\textbf{Hardware Suitability} 
    & Neuromorphic processors, FPGAs 
    & GPUs, TPUs, CPUs \\
    \addlinespace[4pt]

\textbf{Data Sparsity Handling} 
    & Naturally exploits input and activation sparsity 
    & Less efficient with sparse data unless specifically designed \\
    \addlinespace[4pt]

\textbf{Noise Robustness} 
    & Potentially higher due to temporal dynamics and population coding 
    & Can be sensitive to noisy inputs \\
    \addlinespace[4pt]

\textbf{Latency} 
    & Potentially very low due to fast spike propagation and sparse events 
    & Can be higher, especially for deep architectures \\
    \addlinespace[4pt]

\textbf{Primary Use Cases (Edge)}
    & Real-time sensor processing, low-power wearables, event-based vision 
    & General AI tasks, often with higher resource availability \\
\bottomrule
\end{tabularx}
\end{table*}

\section{The Edge-SNN Synergy: Architectural Alignment}
\label{sec:edge_suitability}

The deployment of Artificial Intelligence at the edge is not merely a software optimization challenge; it is a physical struggle against the strict limitations of Size, Weight, and Power (SWaP). While traditional approaches attempt to compress continuous-valued Deep Neural Networks (DNNs) to fit these boundaries, Spiking Neural Networks (SNNs) offer a fundamental divergence in design philosophy. Rather than being a scaled-down approximation of data-center models, SNNs exhibit an intrinsic architectural alignment with the sporadic, event-driven nature of the physical world. This section deconstructs this synergy across three critical dimensions: the physics of SWaP, the demands of always-on sensing, and the elimination of synchronous overhead in the processing pipeline.

\subsection{The SWaP Imperative: Beyond Just Energy}
\label{subsec:swap_imperative}

\begin{figure}[htb]
    \centering
    \includegraphics[width=1.0\columnwidth]{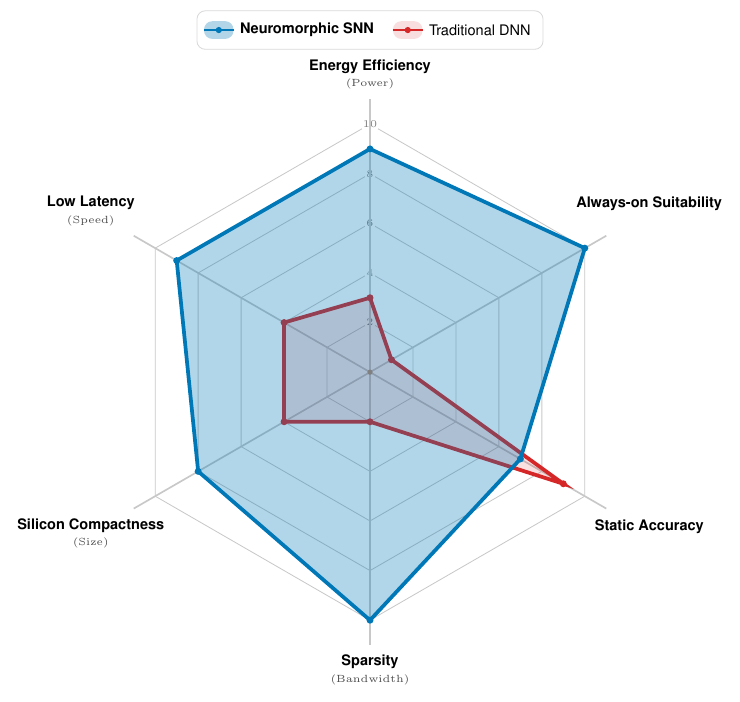}
    \caption{\textbf{Radar chart illustrating the multidimensional trade-offs between Neuromorphic SNNs and a traditional Edge DNN baseline.} While the DNN maintains a marginal advantage in static accuracy, the SNN demonstrates superior performance in SWaP-constrained metrics—specifically Energy Efficiency, Sparsity, and Always-on Suitability. This architectural alignment identifies SNNs as the optimal candidate for resource-limited edge deployment despite the accuracy trade-off.}
    \label{fig:swap_radar}
\end{figure}

The constraints of edge computing are multidimensional. The suitability of SNNs must be evaluated not only through the lens of energy consumption but through the interconnected triad of Size (Silicon Area), Weight (Battery Constraints), and Power (Thermal Design). As illustrated in Fig.~\ref{fig:swap_radar}, this multidimensional comparison reveals that SNNs offer a favorable architectural alignment for constrained environments, balancing a marginal drop in accuracy with significant gains in efficiency metrics.

\subsubsection{Power and Thermal Constraints}
In conventional CMOS scaling, the industry faces the ``Dark Silicon'' phenomenon, where power density limits prevent all transistors on a chip from operating simultaneously without inducing thermal throttling \cite{esmaeilzadeh2011dark, taylor2012dark, hardavellas2011toward, shafique2014eda}. Standard DNN accelerators, which rely on dense matrix multiplications, risk violating these Thermal Design Power (TDP) envelopes. SNNs mitigate this through \textbf{activation sparsity}. By processing information only when salient events occur, SNNs effectively ``light up'' dark silicon only on demand, maintaining high functional density without breaching thermal limits \cite{Iaboni2024, roy2019towards}.

\subsubsection{Weight and Mobility}
For mobile edge agents, such as micro-UAVs or wearable health monitors, the primary constraint is often weight—specifically, the battery mass required for sustained operation. There is a direct correlation between algorithmic efficiency and flight time or operational lifespan. By reducing the inference energy cost by orders of magnitude (often exceeding $100\times$ reduction in neuromorphic implementations \cite{davies2021advancing, rathi2023diet}), SNNs allow for significantly smaller battery payloads. This reduction in energy storage requirements directly translates to enhanced mobility and longer mission durations for weight-critical autonomous systems.

\subsubsection{Size and Silicon Real Estate}
SNNs facilitate compact physical implementations. The binary nature of spikes eliminates the need for complex floating-point units (FPUs) and large multipliers. Instead, synaptic operations can be reduced to simple accumulations (AC), drastically shrinking the logic gate count and silicon area required per neuron compared to traditional Multiply-Accumulate (MAC) units used in DNNs.

\subsection{The ``Always-On'' Sensing Paradigm}
\label{subsec:always_on}

A quintessential edge workload is the ``always-on'' monitor—systems that must continuously analyze the environment for specific triggers (e.g., keyword spotting, seismic anomaly detection, or security surveillance) while remaining in a deep sleep state.

\subsubsection{The Zero-Input, Zero-Energy Advantage}
DNNs suffer from high static power consumption in these scenarios. Even when processing silence or a static image, a traditional Convolutional Neural Network (CNN) executes a fixed number of operations per frame to confirm the absence of a target. Conversely, SNNs offer a \textbf{``Zero-input, Zero-energy''} characteristic. In the absence of sensory change, no spikes are generated, and the dynamic power consumption drops to near zero.

\subsubsection{Temporal Sparsity}
This makes SNNs uniquely suited for wake-up controllers. The network remains quiescent, consuming negligible power, until the membrane potential of input neurons accumulates sufficient charge from a valid signal to cross the firing threshold \cite{Gerstner2002}. This event-driven wake-up mechanism stands in stark contrast to the polling-based architecture of conventional AI, ensuring that energy is expended only when information is actually present.

\subsection{The Sensor-Algorithm-Hardware Pipeline: End-to-End Asynchrony}
\label{subsec:pipeline}

The efficiency of SNNs is fully realized only when the algorithm is part of a holistic, asynchronous pipeline that spans from the sensor to the silicon substrate.

\subsubsection{Seamless Sensory Integration}
SNNs interface naturally with event-based sensors, such as Dynamic Vision Sensors (DVS) \cite{gallego2022event, Ahmed2025}. Unlike frame-based cameras that flood the processor with redundant pixel data at fixed intervals (e.g., 60 Hz), DVS pixels operate asynchronously, outputting data only upon detecting intensity changes. This creates a ``spike-in, spike-out'' data path, avoiding the computationally expensive transcoding of sparse events into dense frames required by traditional ANNs.

\subsubsection{Eliminating the Clock Overhead}
Perhaps the most profound advantage is the potential for \textbf{End-to-End Asynchrony}. In traditional digital design, the global clock distribution network can consume up to 30--40\% of a chip's dynamic power \cite{horowitz20141, rabaey2002digital}. SNNs, particularly when implemented on neuromorphic substrates like Intel Loihi 2 or asynchronous FPGA architectures \cite{karamimanesh2025fpga, modha2023neural}, do not require global synchronization. They operate on local handshakes. This eliminates the ``von Neumann bottleneck'' of waiting for memory fetches in lockstep with a clock cycle, allowing the hardware to mirror the continuous, asynchronous dynamics of the physical world.

\vspace{0.5em}
\noindent \textit{The transition from DNNs to SNNs at the edge represents more than a gain in efficiency; it marks a paradigm shift from ``High-Precision Computing''—where the goal is exact numerical reconstruction of the input—to ``Actionable Intelligence.'' By discarding redundant temporal data and prioritizing salient events, SNNs align the computational cost directly with the complexity of the action required, making them the critical enabler for pervasive, autonomous edge intelligence.}
\section{Recent Advancements in Algorithms and Optimization}
\label{sec:advancements}

\begin{figure*}[t]
    \centering
    \includegraphics[width=0.95\linewidth]{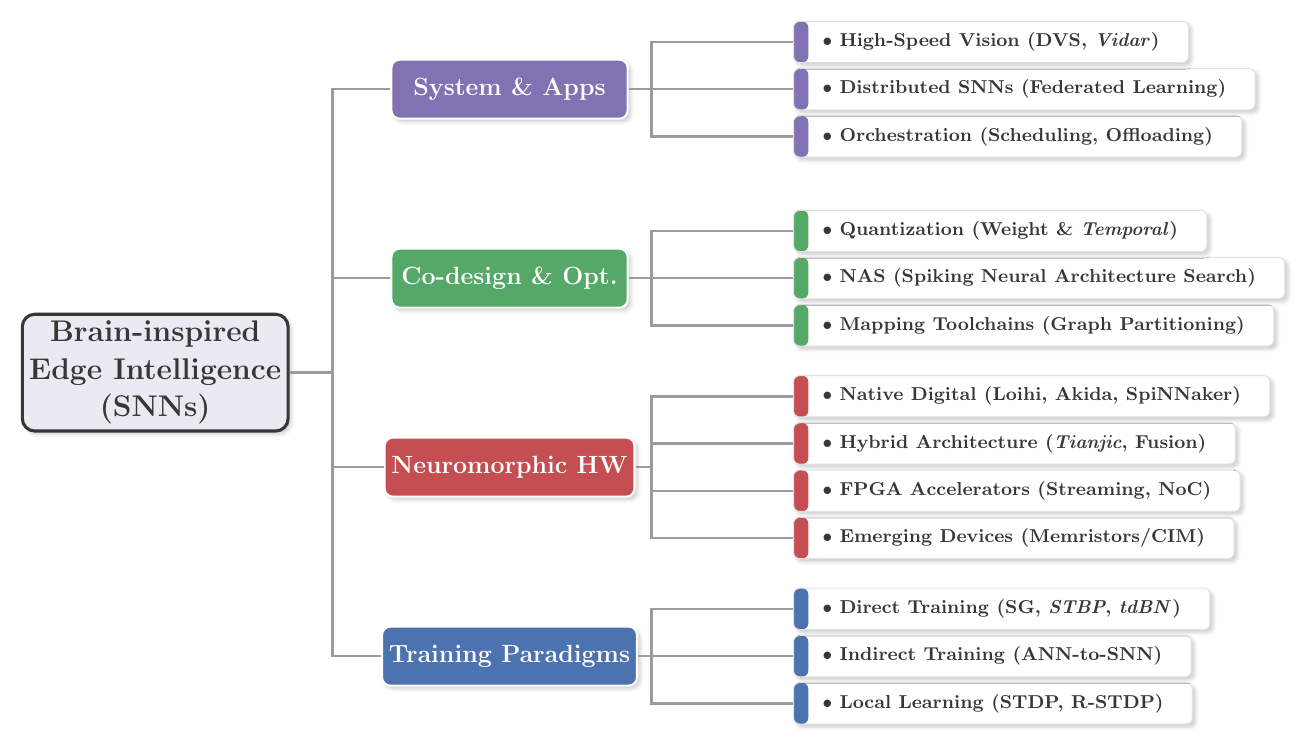}
    \caption{\textbf{Taxonomy of Brain-inspired Edge Intelligence.} The survey is structured around four pillars: 
(1) \textbf{Training Paradigms}: Highlighting the shift from conversion to direct training algorithms like \textbf{STBP} and \textbf{tdBN}; 
(2) \textbf{Neuromorphic Hardware}: Covering native digital chips, \textbf{FPGA} accelerators, and hybrid architectures like \textbf{Tianjic}; 
(3) \textbf{Co-design Strategies}: Focusing on \textbf{NAS} and temporal quantization; 
(4) \textbf{System \& Applications}: Emphasizing high-speed vision (\textbf{Vidar}) and edge-cloud orchestration.}
    \label{fig:taxonomy}
\end{figure*}

The landscape of SNNs for edge computing has evolved rapidly from theoretical exploration to system-level deployment. Moving beyond simple conversion methods, the field has witnessed a paradigm shift towards hardware-aware direct training, heterogeneous architecture design, and multi-dimensional co-optimization. To systematically navigate these diverse developments, we structure our review around the taxonomy presented in Fig.~\ref{fig:taxonomy}, categorizing the field into four pillars: training paradigms, optimization strategies, neuromorphic hardware, and system integration.

\subsection{Training Paradigms: From Gradient Approximation to Fast Conversion}
\label{sec:training}

The training landscape of SNNs has evolved from biological mimicry to high-performance computing. We categorize recent advancements into two distinct eras: the \textit{Gradient-based Era}, which solves the non-differentiability of spikes for direct learning, and the \textit{Low-Latency Conversion Era}, which redefines the speed-accuracy trade-off.

\subsubsection{Gradient-based Era: Surmounting Non-Differentiability}
The fundamental obstacle in training deep SNNs is the non-differentiable nature of the Heaviside step function used for spike generation, which blocks gradient flow during backpropagation.
To address this, the \textbf{Surrogate Gradient (SG)} methodology was proposed to approximate the derivative of the spike function with a continuous pseudo-derivative (e.g., sigmoid or triangular shapes) during the backward pass \cite{zenke2018superspike, neftci2019surrogate, li2021differentiable, Eshraghian2023Survey}.
Pioneering this direction, the \textbf{Spatio-Temporal Backpropagation (STBP)} algorithm \cite{Wu2018STBP, Wu2019STBP_PAMI} was introduced to explicitly aggregate gradients along both spatial and temporal dimensions. Unlike traditional BPTT, which often neglects spatial layer-wise dependencies, STBP enables the training of high-performance SNNs directly on complex datasets and has become the standard for modern SNN optimization.

However, scaling SG-based training to deep architectures (e.g., ResNet-50) introduced a secondary challenge: gradient vanishing or exploding due to the binary nature of spike propagation.
To tackle this, \textbf{Threshold-dependent Batch Normalization (tdBN)} \cite{Zheng2021tdBN} was proposed. By normalizing the membrane potential along the temporal dimension with a threshold-aware scaling factor, tdBN ensures that neurons maintain a healthy firing rate, effectively regulating gradient magnitude.
Further refining the learning dynamics, recent works have moved beyond fixed surrogate shapes. The Dspike framework \cite{li2021differentiable} allows the steepness of the surrogate gradient to be learned adaptively, dynamically balancing the trade-off between accurate gradient estimation and training stability.

Addressing the high memory cost of unrolling SNNs over time in SG methods, research is pivoting towards \textbf{Event-Driven Learning}. To bridge the gap between high-performance backpropagation and always-on edge constraints, \textbf{Spike-Timing-Dependent Event-Driven (STD-ED)} algorithms \cite{WeiWenjie2024} have been developed. These methods trigger weight updates only upon spike events rather than at every global timestep, reducing energy consumption by approximately $30\times$.

\subsubsection{Low-Latency Conversion Era: Breaking the Time Barrier}
While direct training offers temporal efficiency, it often lags behind ANNs in absolute accuracy on large-scale tasks. ANN-to-SNN conversion traditionally bridged this accuracy gap but suffered from a severe limitation: \textbf{extreme inference latency} (hundreds of timesteps) required to approximate continuous activation values via rate coding.
To mitigate this latency bottleneck, recent efforts have shifted focus from simple firing rate approximation to temporal information compression \cite{ding2021optimal, bu2025inference, song2024one}. Techniques such as calibrating forward temporal bias and introducing burst spikes have been employed to shorten the inference window.
Taking a more radical approach to eliminate the iterative nature of conversion, a \textbf{parallel conversion learning framework} \cite{HaoZecheng2024} was recently introduced. By establishing a mapping between parallel neuron timesteps and cumulative spike rates, this method achieves 72.90\% top-1 accuracy on ImageNet-1k (ResNet-34) in merely \textbf{4 timesteps}, marking a milestone where conversion-based methods finally rival the low-latency advantages of direct training.

\vspace{0.5em}
\noindent \textit{A convergence of paradigms is emerging. The strict dichotomy between ``Direct Training'' and ``Conversion'' is dissolving into \textbf{Hybrid Training} strategies. Recent frameworks \cite{YuDi2025, Seekings2024} demonstrate that SNNs are no longer solitary models but are increasingly deployed as ``Neural Front-ends'' for feature extraction, coupled with ANN back-ends for semantic classification, or optimized via ANN-guided knowledge distillation. Future training algorithms will likely prioritize this heterogeneous synergy over pure standalone performance.}

\subsection{Optimization: Hardware-Aware Model Compression}
\label{sec:optimization}

Deploying SNNs on resource-constrained edge devices requires navigating a complex trade-off between model accuracy, memory footprint, and energy consumption. Recent research has unified these objectives under hardware-aware compression, targeting three critical dimensions: connectivity (Pruning), topology (NAS), and precision (Quantization).

\subsubsection{Pruning: Minimizing Redundant Connectivity}
Standard SNNs often inherit dense connectivity patterns that are computationally redundant for sparse spiking data.
To address this, \textbf{Adaptive Pruning} strategies have been developed to dynamically adjust pruning rates based on neural activity. For instance, \textbf{dynamic rollback mechanisms} \cite{Rivelli2025} have been utilized to achieve sub-$\mu$W power consumption on neuromorphic hardware.
However, unstructured pruning often leads to irregular memory access patterns that are difficult to accelerate. Consequently, recent works like the \textbf{Optimal Spiking Brain Compression (OSBC)} \cite{ShiLianfeng2025} have adapted Optimal Brain Compression (OBC) theory to SNNs. By minimizing the loss on the membrane potential rather than just weights, OSBC achieves 97\% sparsity with negligible accuracy loss (e.g., $<2\%$ drop on DVS128-Gesture), proving that SNN redundancy is significantly higher than that of ANNs.

\subsubsection{Neural Architecture Search (NAS): Automating Topology Design}
Manual heuristic design of SNNs (e.g., simply copying VGG or ResNet) often fails to capture the intrinsic temporal dynamics of spikes.
To overcome this human bias, \textbf{Neural Architecture Search (NAS)} has been introduced to automate the discovery of spike-friendly topologies \cite{svoboda2025spiking, Na2022AutoSNN}.
Early approaches like SNAS \cite{kim2022neural} relaxed the discrete search space into a continuous domain. However, a unique challenge in SNN-NAS is the mismatch between topology search and the surrogate gradients used for training.
Addressing this, \textbf{SpikeDHS} \cite{Che2022SpikeDHS} proposed a joint optimization framework that simultaneously searches for the network topology and the optimal shape of the surrogate gradient.
Furthermore, as deployment targets diversify, the field has evolved towards \textbf{Hardware-Aware NAS}. Recent neuroevolutionary approaches \cite{Yik2024Neuroevolution} explicitly incorporate hardware metrics—such as latency and energy—into the search objective, ensuring the discovered architectures are located on the Pareto-optimal frontier of the target device.

\subsubsection{Quantization: Compressing Precision and Time}
While traditional quantization reduces weight bit-width, optimizing SNNs offers a unique additional dimension: \textbf{temporal quantization} (reducing simulation timesteps $T$).
Reducing $T$ provides linear gains in latency but typically degrades accuracy. To resolve this, the \textbf{QUEST} framework \cite{LiSai2025} introduced a co-design approach involving adaptive precision and quantization-aware training. By unifying low-bit weights (2-bit) with reduced timesteps, QUEST achieves a $\sim$93x energy improvement over equivalent ANNs.
Similarly, for autonomous agents where response time is critical, \textbf{SNN4Agents} \cite{PutraRachmad2024} demonstrates that jointly quantizing weights (to 10-bit) and reducing attention windows can yield a $4\times$ efficiency gain, validating that multi-dimensional quantization is essential for real-time edge intelligence.

\vspace{0.5em}
\noindent \textit{The trajectory of SNN optimization is shifting from single-variable compression (weight pruning only) to \textbf{Multi-Dimensional Co-Optimization}. Future frameworks will likely treat weight sparsity, bit precision, and temporal resolution (timesteps) as coupled variables, optimizing them jointly to align with the specific constraints of non-Von Neumann neuromorphic hardware.}

\subsection{Neuromorphic Hardware: Evolution of Computational Substrates}
\label{sec:hardware}

\begin{table*}[t]
    \centering
    \caption{Prominent Neuromorphic Hardware Platforms Relevant for Edge SNNs (2023-2025 Updates)}
    \label{tab:neuromorphic_platforms}
    
    \small 
    \renewcommand{\arraystretch}{1.3} 
    
    \begin{tabularx}{\linewidth}{@{} l l >{\hsize=1.4\hsize}X l c c c >{\hsize=0.6\hsize}X c @{}}
    \toprule 
    \textbf{Platform} & \textbf{Dev.} & \textbf{Key Features for Edge SNNs} & \textbf{Capacity} & \textbf{Learn.} & \textbf{Eff.} & \textbf{Scale} & \textbf{Edge Relevance} & \textbf{Ref.} \\
    \midrule 
    
    \textbf{Loihi 2} & Intel 
        & Programmable spiking neurons, dynamic synapses, async processing 
        & $>$1 M & Yes & \CIRCLE & \CIRCLE 
        & Robotics, AI control & \cite{davies2021advancing} \\
    \addlinespace[4pt]

    \textbf{Hala Point} & Intel 
        & System of 1,152 Loihi 2 chips, massive parallelism 
        & 1.15 B & Yes & \LEFTcircle & \CIRCLE\textbf{+} 
        & Large-scale AI research & \cite{IntelHalaPoint} \\
    \addlinespace[4pt]

    \textbf{Darwin3} & ZJU 
        & Rack-scale system, specialized SNN ISA, DarwinOS support 
        & $>$2 B & Yes & \LEFTcircle & \CIRCLE\textbf{+} 
        & Brain simulation, AGI & \cite{Ma2024Darwin3} \\
    \addlinespace[4pt]

    \textbf{NorthPole} & IBM 
        & In-memory computing, inference-only optimization 
        & N/A & No & \CIRCLE\textbf{+} & N/A 
        & Ultra-low-power AI & \cite{modha2023neural} \\
    \addlinespace[4pt]

    \textbf{Akida 2} & BrainChip 
        & Event-based, IP for SoC integration, TENNs support
        & Scalable & Yes & \CIRCLE\textbf{+} & \CIRCLE 
        & IoT, wearables, smart home & \cite{BrainChipAkida} \\
    \addlinespace[4pt]

    \textbf{SpiNNaker2} & Manch.
        & ARM cores, custom spike routing 
        & $>$1 M/bd & Soft. & \Circle & \CIRCLE\textbf{+} 
        & Robotics, embedded sys. & \cite{mayr2019spinnaker} \\
    \addlinespace[4pt]

    \textbf{Tianjic} & Tsinghua 
        & Hybrid ANN-SNN architecture 
        & $10^5$ & Yes & \LEFTcircle & \CIRCLE 
        & Autonomous driving & \cite{Pei2019Tianjic} \\
    
    \bottomrule 
    \end{tabularx}

    \vspace{1mm}
    \parbox{\linewidth}{\footnotesize 
    \textbf{Legend:} \textbf{Eff.} = Power Efficiency; \textbf{Scale} = Scalability. \\
    Ratings: \CIRCLE\textbf{+} = Best (Ultra Low Power/High Scale); \CIRCLE = Very High; \LEFTcircle = High; \Circle = Moderate. \\
    \textit{Note:} M: Million neurons, B: Billion neurons. Capacity varies by specific implementation.
    }
\end{table*}

Deploying SNNs at the edge necessitates hardware that transcends the energy-latency trade-offs inherent in the von Neumann bottleneck. The evolution of neuromorphic hardware has transitioned from simple biological mimicry to sophisticated, domain-specific architectures. We categorize recent advancements into four distinct architectural philosophies: Digital Asynchronous primitives, Hybrid paradigms, Scale-out Systems, and FPGA-based reconfigurable accelerators. Table \ref{tab:neuromorphic_platforms} summarizes these platforms, contrasting their capacity, learning plasticity, and suitability for edge deployment against standard commercial-off-the-shelf (COTS) hardware.

\subsubsection{Digital Asynchronous Architectures: The Efficiency Baseline}
The foundational philosophy of neuromorphic engineering prioritizes energy efficiency through event-driven asynchronous communication. Prominent examples, such as \textbf{Intel Loihi} and \textbf{IBM TrueNorth}, utilize fine-grained power gating where cores remain idle until a spike event occurs \cite{orchard2021efficient, merolla2014million, akopyan2015truenorth}.
\textbf{Contrast with Edge GPUs:} Unlike edge GPUs (e.g., NVIDIA Jetson), which rely on batch processing and continuous memory access (high static power), asynchronous architectures achieve orders-of-magnitude lower dynamic power consumption for sparse workloads. However, their rigid adherence to SNN dynamics often limits their ability to support standard DNN layers effectively, necessitating the shift toward hybrid designs.

\subsubsection{Hybrid Architectures: Cross-Paradigm Fusion}
To address the rigidity of pure neuromorphic designs, \textbf{Hybrid Architectures} have emerged, aiming to bridge Computer Science (CS) accuracy with Neuroscience (NS) efficiency. A seminal breakthrough is the \textbf{Tianjic} chip \cite{Pei2019Tianjic}, which introduces a unified functional core (FHC) capable of concurrent multi-mode operation. By integrating configurable unified buffers and adaptable axons, Tianjic allows diverse models---from biologically plausible SNNs to standard CNNs---to coexist on the same substrate.
Following this paradigm, several emergent designs have adopted this heterogeneous strategy to support ANN-SNN conversion natively on-chip \cite{mayr2019spinnaker, yao2020fully, zhao2022spiking, christensen20222022}.
\textbf{Contrast with Edge CPUs:} While embedded CPUs offer general-purpose flexibility, they lack the massive parallelism required for real-time sensor fusion. Hybrid chips fill this gap, enabling systems like autonomous robots to utilize SNNs for low-latency reflex (obstacle avoidance) and ANNs for high-level cognition (semantic recognition) simultaneously.

\subsubsection{Scale-out Systems: From Chip to Brain-Scale}
While edge inference focuses on single-chip efficiency, a parallel track of research addresses the scalability challenges of whole-brain simulation. The \textbf{Darwin} series \cite{Ma2017Darwin} represents a significant effort in this direction. The evolution from early prototypes to the massive \textbf{Darwin3} chip \cite{Ma2024Darwin3} reflects a specialized Instruction Set Architecture (ISA) optimized for spiking dynamics and on-chip plasticity.
Crucially, the focus has shifted to system-level integration. The \textbf{Darwin Monkey (``Wukong'')} system integrates nearly 1,000 Darwin3 chips via high-speed distinct interconnects, achieving a scale of over 2 billion neurons and 100 billion synapses. This surpasses many existing platforms, including recent Intel Hala Point configurations, by focusing on addressing the interconnect bottlenecks inherent in massive-scale neuromorphic clusters \cite{davies2018loihi, furber2014spinnaker, pehle2022brainscales, chen2022asurvey}.
\textbf{Contrast with Data Center GPUs:} Unlike GPU clusters that suffer from communication overhead during sparse spike exchange, Darwin Monkey utilizes specialized routing to support brain-inspired operating systems (DarwinOS) efficiently.

\subsubsection{FPGA-based Accelerators: The Pragmatic Middle Ground}
Field-Programmable Gate Arrays (FPGAs) have emerged as a vital ``middle ground,'' offering the reconfigurability absent in ASICs and the deterministic latency lacking in GPUs. Recent research focuses on overcoming the unique mapping challenges of SNNs on synchronous fabrics \cite{karamimanesh2025fpga}.

\par \textbf{Breaking the Memory Wall:}
SNNs require persistent state storage (membrane potentials, $V_{mem}$), creating significant bandwidth pressure. To mitigate this, the \textbf{S2N2} streaming architecture \cite{Li2021S2N2} was proposed to optimize BRAM usage by restricting state updates to active neurons. Challenging the event-driven dogma, \textbf{SyncNN} \cite{Panchapakesan2021} demonstrated that a \textbf{synchronous dataflow} on FPGAs can actually outperform asynchronous mapping, improving throughput by \textbf{4.6$\times$} and energy efficiency by \textbf{2.3$\times$} compared to state-of-the-art event-driven accelerators.
Similarly, the Weight-Stationary Local-Output-Stationary (WS-LOS) dataflow \cite{Kang2023TCAS} was developed to maximize data reuse, achieving ultra-low energy consumption (24.3 $\mu$J/Image).

\par \textbf{Sparsity-Aware Pipelines:}
Addressing the irregular sparsity of spikes, \textbf{FireFly v2} \cite{FireFlyV2_2024} introduces a spatiotemporal dataflow that eliminates storage conflicts, achieving a clock frequency of 600 MHz. Furthermore, \textbf{SpikeX} \cite{SpikeX2025} proposes a hardware-software co-optimization framework for unstructured sparsity. By dynamically rebalancing systolic arrays, SpikeX reduces the Energy-Delay Product (EDP) by over $15\times$ compared to dense baselines.

\vspace{0.5em}
\noindent \textit{The trajectory of neuromorphic hardware is undergoing a fundamental shift from ``strict biological mimicry'' to ``functional heterogeneity.'' Early designs prioritized the faithful replication of neuronal dynamics, often at the cost of computational flexibility. Current trends, however, favor software-defined architectures where hardware resources can be dynamically reallocated between synchronous (ANN-like) and asynchronous (SNN-like) modes. This evolution suggests that the future of edge intelligence lies not in replacing von Neumann architectures entirely, but in augmenting them with domain-specific, event-driven accelerators that can seamlessly integrate into standard computing stacks.}

\subsection{The Software Gap: Toolchains and Compilers}
\label{subsec:software_gap}

While neuromorphic hardware has seen rapid acceleration, the software ecosystem remains the primary bottleneck hindering widespread adoption. Unlike the mature CUDA ecosystem for GPUs, the SNN landscape currently suffers from severe \textit{fragmentation}. Developers are often forced to manually optimize spiking dynamics for specific hardware architectures due to the absence of a unified \textbf{Intermediate Representation (IR)} and a standardized compiler stack.

\subsubsection{Mapping Toolchains: Solving the Embedding Problem}
The translation of a logical SNN graph (neurons and synapses) onto a physical neuromorphic fabric (cores and Network-on-Chip routers) is a non-trivial combinatorial optimization problem known as the ``Mapping Problem'' \cite{balaji2019mapping, urgese2016optimizing, schuman2022opportunities}. Recent advancements focus on bridging this gap through automated toolchains that balance communication bandwidth, energy consumption, and latency.

\par \textbf{NeuMap} addresses the challenge of mapping SNNs to multicore architectures by treating the network partition as a cluster-to-core allocation problem. By analyzing spike communication patterns and hardware constraints, it employs meta-heuristics to optimize placement, achieving an energy reduction of up to 84\% and latency improvements of 55\% compared to prior baselines like SpiNeMap.

\par \textbf{EdgeMap} extends this logic specifically for resource-constrained edge devices \cite{XueJianwei2023}. It introduces a streaming-based partitioning method that rigorously accounts for fan-in/fan-out limits. Utilizing NSGA-II-based multi-objective optimization, EdgeMap navigates the trade-off between energy and communication costs. Benchmark results highlight its efficacy, demonstrating a 1225x enhancement in execution time and a 57\% reduction in energy compared to conventional mapping schemes.

\subsubsection{Software Frameworks and Fragmentation}
The current development environment is bifurcated into vendor-specific toolchains and general-purpose libraries, as summarized in Table \ref{tab:snn_software}.

\begin{table*}[t]
    \centering
    \caption{Selected SNN Software Frameworks and Mapping Toolchains (2020--2025)}
    \label{tab:snn_software}
    \small 
    \renewcommand{\arraystretch}{1.3} 
    
    \begin{tabularx}{\linewidth}{@{} l >{\hsize=1.1\hsize}X >{\hsize=0.9\hsize}X c @{}}
    \toprule 
    \textbf{Framework (Developer)} & \textbf{Core Features \& Purpose} & \textbf{Edge-Relevant Capabilities} & \textbf{Ref.} \\
    \midrule 
    
    \multicolumn{4}{l}{\textit{\textbf{Hardware Mapping Toolchains}}} \\
    \addlinespace[3pt]
    
    \textbf{NeuMap} (NUDT) 
        & Mapping SNNs to multicore HW; Comm. pattern calculation; Meta-heuristic cluster-to-core mapping 
        & Optimized for NoC latency and energy on multicore edge hardware 
        & \cite{XiaoChao2022} \\
    \addlinespace[4pt]
    
    \textbf{EdgeMap} (SJTU) 
        & Optimized mapping for edge; Streaming-based partitioning; Multi-objective optimization (NSGA-II)
        & Significant reduction in latency, energy, and comm. cost for edge deployment 
        & \cite{XueJianwei2023} \\
    \addlinespace[8pt] 
    
    \multicolumn{4}{l}{\textit{\textbf{Deep Learning \& Deployment Frameworks}}} \\
    \addlinespace[3pt]

    \textbf{Lava} (Intel) 
        & Neuro-inspired app dev; Modular; Platform-agnostic; Supports deep learning (lava-dl) 
        & Native deployment on Loihi chips; Optimized for energy/speed efficiency 
        & \cite{GomeWalter2023} \\
    \addlinespace[4pt]

    \textbf{SNNTorch} (UCSC) 
        & PyTorch-based training; Surrogate gradients; Temporal dynamics; GPU acceleration 
        & Facilitates training of deep SNNs deployable on edge; Seamless PyTorch integration 
        & \cite{Eshraghian2023Survey} \\
    \addlinespace[4pt]
    
    \textbf{SpikingJelly} (PKU) 
        & PyTorch-based; LIF/PLIF neurons; Direct training (SG); Rich encoding methods 
        & Enables development of deep SNNs suitable for direct edge deployment 
        & \cite{HuoBingqiang2025} \\
    \addlinespace[8pt]

    \multicolumn{4}{l}{\textit{\textbf{Bio-inspired Simulation \& Prototyping}}} \\
    \addlinespace[3pt]

    \textbf{Nengo} (ABR) 
        & Large-scale modelling (NEF); Supports multiple backends (Loihi, SpiNNaker, CPU/GPU) 
        & Real-time control applications verified on diverse neuromorphic platforms 
        & \cite{Bekolay2014} \\
    \addlinespace[4pt]
    
    \textbf{BindsNET} (UMass) 
        & Bio-plausible simulation; STDP/R-STDP rules; Event-driven dynamics 
        & Prototyping SNNs with on-device online learning potential 
        & \cite{Hazan2018} \\
    \addlinespace[4pt]
    
    \textbf{Brian2} (Sorbonne) 
        & Biophysical simulation; Flexible model definition; Code generation 
        & Simulating complex, biologically detailed SNNs prior to hardware mapping 
        & \cite{Stimberg2019} \\
        
    \bottomrule 
    \end{tabularx}
\end{table*}

\par \textbf{Vendor-Specific Ecosystems:} Intel's \textbf{Lava} represents a shift towards modularity, offering an open-source framework for Loihi and other neuro-inspired processors. It supports hyper-granular parallelism and dynamic neural fields (DNF), yet its full potential is tightly coupled with the Loihi architecture \cite{GomeWalter2023}. Similarly, frameworks like MetaTF remain bound to Akida hardware.

\par \textbf{General-Purpose Libraries:} On the algorithmic front, PyTorch-based libraries such as \textbf{SpikingJelly} and \textbf{SNNTorch} have lowered the entry barrier for researchers by integrating SNN dynamics into standard deep learning workflows \cite{ZhouChenlin2024}. Other established simulators like Brian2 and Nengo continue to serve the computational neuroscience community \cite{kulkarni2021benchmarking, manna2023frameworks, al2024survey}. 
To navigate this fragmented ecosystem, a comprehensive multimodal benchmark \cite{cheng2025comprehensive} has recently provided a quantitative evaluation of frameworks including SpikingJelly, BrainCog, and Lava. This study highlights critical trade-offs in energy efficiency and latency across diverse tasks, offering a data-driven basis for selecting appropriate toolchains for edge deployment.
Crucially, these tools primarily address intra-node algorithmic fidelity. To bridge the gap towards realistic deployment, system-level simulators like \textbf{iFogSim} \cite{gupta2017ifogsim} are increasingly essential for modeling the inter-node latency, bandwidth limits, and energy consumption inherent in distributed edge environments.
However, the lack of a seamless compilation path from these high-level Python libraries to heterogeneous neuromorphic chips remains a critical ``Software Gap.''

\subsection{System-Level Integration: Cloud, Edge, and IoT}
\label{subsec:system_integration}

The maturation of SNN algorithms has enabled a transition from standalone acceleration to \textbf{Distributed Edge Intelligence}. This paradigm shifts focus from single-chip inference to collaborative systems where SNNs operate within broader IoT and cloud ecosystems.

\par \textbf{Distributed Wireless Intelligence:} Deploying SNNs across distributed sensor nodes introduces unique challenges in bandwidth and synchronization. Recent work models this as a joint optimization problem, minimizing energy under strict spike-loss constraints over wireless channels \cite{LiuYanzhen2024}. This ensures that distributed SNNs can maintain high inference accuracy even in noisy IoT environments.

\par
\begin{figure*}[t]
    \centering
    \includegraphics[width=0.85\linewidth]{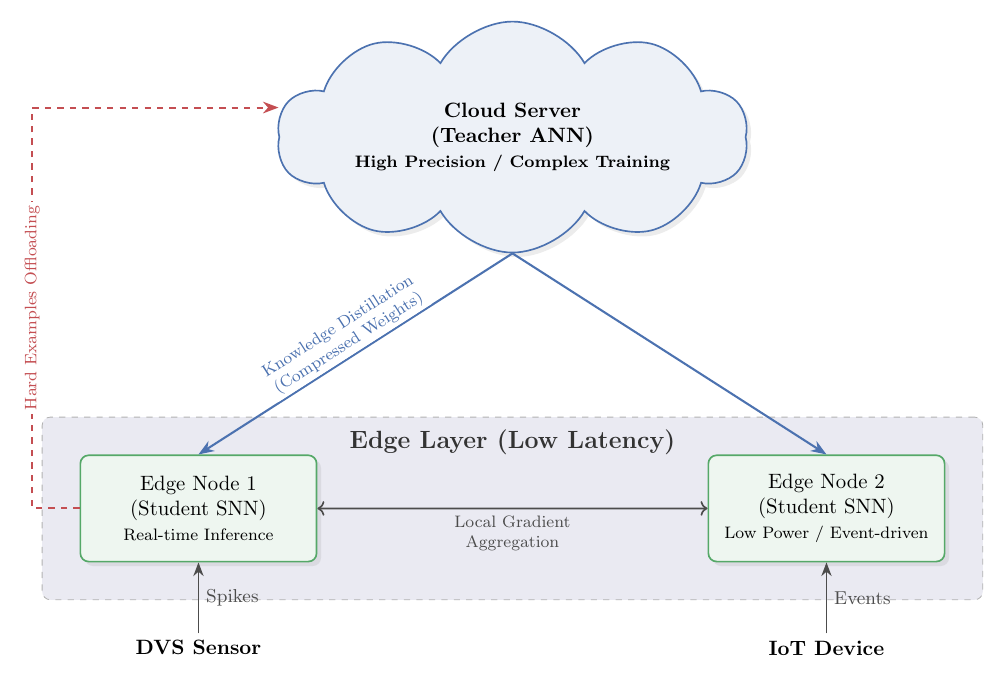}

    \caption{\textbf{Proposed SNN-based Cloud-Edge Collaboration Framework.} The cloud server hosts a complex Teacher ANN to guide lightweight Student SNNs on edge nodes via knowledge distillation. Edge nodes process real-time event streams and only offload low-confidence ``hard examples'' to the cloud, significantly reducing bandwidth usage.}
    \label{fig:edge_cloud_snn}
\end{figure*}

 \textbf{Edge-Cloud Collaboration:} To balance the limited capacity of edge devices with the computational power of the cloud, hierarchical architectures have emerged (Fig.~\ref{fig:edge_cloud_snn}). To facilitate such hierarchical collaboration, lightweight integration frameworks like \textbf{FogBus} \cite{tuli2019fogbus} have been proposed to manage the heterogeneity and data flow between resource-constrained edge nodes and the cloud. Building on this paradigm, frameworks like \textbf{ECC-SNN} leverage \textit{knowledge distillation}, where a complex cloud-based ANN acts as a ``teacher'' guiding lightweight edge SNN ``students'' \cite{YuDi2025}. A key innovation here is the confidence-based offloading mechanism: edge nodes process real-time event streams locally, transmitting only low-confidence ``hard examples'' to the cloud for high-precision analysis. This strategy significantly reduces communication overhead while preserving global system accuracy.

\par \textbf{Federated Learning for SNNs:} Addressing data privacy and the ``data island'' problem, Federated Learning (FL) has been adapted for spiking domains. Techniques such as \textbf{FedLEC} tackle the issue of non-IID (label-skewed) data in neuromorphic networks. By employing intra-client label weight calibration and inter-client distillation, these methods mitigate model drift, allowing SNNs to learn collaboratively without sharing raw event data \cite{YuDongfang2024, abbasihafshejani2025spiking, aksu2025privacy, nguyen2025sparsified}.

\vspace{0.5em}
\noindent \textit{The trajectory of SNN software is approaching an inflection point similar to the early days of GPU computing. The current fragmentation—characterized by isolated vendor toolchains and disconnected mapping algorithms—is unsustainable. The future lies in \textbf{Standardization}. The community must converge towards a unified neuromorphic Intermediate Representation (IR) and a cross-platform compiler stack (analogous to LLVM or TVM). Only when a high-level description in PyTorch can be seamlessly compiled to any underlying neuromorphic substrate—be it Loihi, TrueNorth, or FPGA-based accelerators—will SNNs achieve the ubiquity necessary for next-generation ubiquitous computing.}

\section{Applications}
\label{sec:applications}
\begin{table*}[t]
    \centering
    \caption{Exemplary Applications of SNNs in Edge Computing (2022--2025)}
    \label{tab:snn_applications}
    
    \footnotesize 
    \renewcommand{\arraystretch}{1.3} 
    \begin{tabularx}{\linewidth}{@{} l >{\raggedright\hsize=0.6\hsize}X >{\raggedright\hsize=1.0\hsize}X l >{\raggedright\hsize=1.4\hsize}X c @{}}
    \toprule
    \textbf{Domain} & \textbf{Task} & \textbf{Approach / Model} & \textbf{Hardware} & \textbf{Key Performance Metrics} & \textbf{Ref.} \\
    \midrule
    
\textbf{CV} 
    & Neural Rendering 
    & Spiking-NeRF (Hybrid SNN) 
    & GPU/Simulated 
    & \textbf{Quality:} Comparable PSNR; \textbf{Energy:} 2.27$\times$ reduction vs. NeRF 
    & \cite{LiZiwen2024} \\
\addlinespace[4pt]

    & Gesture Recog. 
    & OSBC (One-shot Pruning) 
    & DVS128-Gesture 
    & \textbf{Sparsity:} 97\%; \textbf{Acc Loss:} 1.7\%; \textbf{Method:} Post-training Quant. 
    & \cite{Shi2025} \\ 
\addlinespace[8pt]
    
    \textbf{Robotics} 
        & Visual Place Recog. 
        & Modular SNN Ensembles 
        & Resource Constrained 
        & Compact modules (1.5k neurons); Sequence matching improves R@1 
        & \cite{Hussaini2025} \\
    \addlinespace[4pt]
    
        & Autonomous Agents 
        & SNN4Agents (Quantization) 
        & Event Sensors 
        & \textbf{Acc:} 84.1\%; \textbf{Mem:} $-$68.8\%; \textbf{Eff:} 4.0$\times$ gain 
        & \cite{PutraRachmad2024} \\
    \addlinespace[8pt]
    
    \textbf{IoT} 
        & HAR (Multimodal) 
        & SNN-driven fusion (Event + Skeleton); SNN-Mamba 
        & N/A 
        & High efficiency via sparse multimodal fusion 
        & \cite{ZhengNaichuan2025} \\
    \addlinespace[4pt]
    
        & Keyword Recog. 
        & On-chip learning SNN 
        & Commodity Neuro. 
        & \textbf{Latency:} $<1$\,ms 
        & \cite{PutraRachmad2025} \\
    \addlinespace[8pt]
    
    \textbf{Healthcare} 
        & Neural Decoding 
        & Adaptively Pruned SNNs 
        & SENECA 
        & \textbf{Power:} 0.18\,$\mu$W/step; \textbf{Eff:} $>10\times$ vs. dense SNNs 
        & \cite{Rivelli2025} \\
    \addlinespace[4pt]
    
        & ECG Class. 
        & SNN on FPGA 
        & FPGA / Neuro. 
        & \textbf{Acc:} 98.2\% (MIT-BIH); Ultra-low power matching DNN acc. 
        & \cite{HizemMoez2025} \\
    \addlinespace[8pt]
    
    \textbf{Industrial} 
        & Defect Detection 
        & SNN Model Strengthening 
        & Ind. Datasets 
        & \textbf{Speed:} 11--27$\times$ faster unlearning vs. retraining 
        & \cite{ZhouDeming2023} \\
    \addlinespace[4pt]
    
        & Fault Diagnosis 
        & Deep Spiking ResNet/GNN 
        & Ind. Sensors 
        & Robust spatiotemporal modeling; High efficiency 
        & \cite{WangHuan2023} \\
    \addlinespace[8pt]
    
    \textbf{Edge-Cloud} 
        & Classification 
        & ECC-SNN (Joint Training) 
        & Edge + Cloud 
        & \textbf{Acc:} +4.1\%; \textbf{Energy:} $-$79\%; \textbf{Lat:} $-$39\% 
        & \cite{YuDi2025} \\
    
    \bottomrule
    \end{tabularx}
\end{table*}

SNNs are transitioning from theoretical models to deployable solutions, driven specifically by the rigid constraints of edge computing environments. Rather than merely offering an alternative to ANNs, SNNs address distinct physical and operational bottlenecks: \textbf{extreme power budgets} (battery life), \textbf{communication bandwidth} (saturation from raw data transmission), \textbf{latency} (safety-critical feedback loops), and \textbf{data privacy} (local processing). This section analyzes recent implementations (2023--2025) through the lens of these constraints. Table \ref{tab:snn_applications} provides a quantitative comparison of these applications against DNN baselines in terms of accuracy, energy efficiency, and latency.

\subsection{Computer Vision: High-Speed and Event-Based Sensing}
\label{sec:vision}
Vision systems generate the highest data volume at the edge, creating a massive bandwidth and energy burden. SNNs mitigate this by processing information only when changes occur in the scene.

\subsubsection{High-Speed Retinomorphic Sensing}
Standard frame-based cameras suffer from motion blur and high data redundancy. While Dynamic Vision Sensors (DVS) address this, they often discard static texture information. To bridge this gap, the \textbf{Spike Camera (Vidar)} architecture was introduced \cite{Huang2022Vidar, Dong2024SpikeCam}. By employing a fovea-like sampling mechanism where pixels fire continuous spikes based on luminance intensity, this paradigm achieves ultra-high temporal resolution (up to 40,000 Hz). This allows SNNs to reconstruct high-speed motion (e.g., rapid mechanical rotation) that is invisible to traditional cameras, while significantly compressing the data stream compared to high-speed video \cite{hu2024bitsnns, shi2024towards, sudevan2025underwater}.

\subsubsection{Low-Power Gesture and Activity Recognition}
For battery-powered Human-Machine Interfaces (HMI), the primary constraint is energy consumption. SNNs exploit the spatial sparsity of gesture events to minimize computation. Implementations on neuromorphic hardware have demonstrated power consumption as low as \textbf{178 mW} during active inference \cite{Amir2017}. 
Recent advancements have extended this capability from simple hand gestures to complex \textbf{Human Activity Recognition (HAR)} \cite{SabbellaHemanth2025}. Notably, a February 2025 study proposed a multimodal framework combining event cameras with Spiking Graph Convolutional Networks (SGCN). By processing skeleton data locally, this system avoids the transmission of intrusive raw video feeds, effectively resolving privacy concerns in smart home environments while maintaining high recognition fidelity \cite{ZhengNaichuan2025, ShiLianfeng2025}.

\subsection{Beyond Vision: Auditory and Olfactory Sensing}
\label{sec:non_vision}
Non-visual modalities are naturally sparse. SNNs leverage this property to solve the ``Always-On'' power dilemma.

\subsubsection{Neuromorphic Auditory Processing}
In smart speakers and hearing aids, the wake-word engine must run continuously. SNNs provide a zero-dynamic-power solution during silence. Benchmarking on the Intel Loihi chip has demonstrated that SNNs outperform GPU-based DNNs in Keywords Spotting (KWS) within a milliwatt envelope \cite{blouw2019benchmarking}.
Advancing this bio-inspired approach, recurrent SNN architectures have been utilized to extract fine-grained temporal features, achieving state-of-the-art accuracy \cite{Yin2021Auditory}. Crucially, an April 2025 study reported \textbf{$<$1ms latency} in keyword recognition on commodity neuromorphic processors \cite{PutraRachmad2025}. 
Beyond speech, SNNs are also transforming \textbf{Environmental Sound Classification (ESC)}. By mimicking human auditory pathways, bio-inspired models can classify background scenes with minimal power, enabling the long-term deployment of acoustic sensors in remote areas without frequent battery replacement \cite{BaekSuwhan2024, blouw2019benchmarking, yarga2024neuromorphic}.

\subsubsection{Neuromorphic Olfaction (E-Nose)}
Olfactory data is high-dimensional but sparse, posing a challenge for standard classifiers that require massive training sets. SNNs address the constraint of \textbf{data scarcity}. Mimicking the mammalian olfactory bulb, neural circuits have been implemented to enable \textbf{``One-Shot Learning''} of hazardous chemicals \cite{Imam2020Olfaction}. Unlike DNNs, such systems can learn a new odor signature from a single exposure. Further industrial validation confirmed that SNNs effectively handle sensor drift in gas identification systems, a common failure point for traditional algorithms \cite{Wang2024Gas}.

\subsection{Robotics and Autonomous Systems}
\label{sec:robotics}
In robotics, the trade-off between on-board computational weight and flight time/operation time is critical. SNNs offer a solution via neuromorphic control loops.

\subsubsection{UAVs and Navigation}
For Unmanned Aerial Vehicles (UAVs), processing power directly competes with battery life. The \textbf{SNN4Agents framework} achieved a 4.03x energy efficiency improvement on the NCARS dataset \cite{PutraRachmad2024}. Additionally, Modular SNNs for Visual Place Recognition (VPR) have enabled scalable mapping on resource-constrained drones \cite{Hussaini2025}, providing sub-millisecond reaction times to obstacles—a critical safety feature when ground communication is lost \cite{mompo2024brain, bing2018survey}.

\subsubsection{Edge-Cloud Orchestration}
While edge SNNs handle reflex tasks, complex queries require cloud support. The \textbf{ECC-SNN framework} (May 2025) illustrates this hierarchy, reporting a 4.15\% accuracy improvement on CIFAR-10 by dynamically offloading low-confidence ``hard examples'' to the cloud \cite{YuDi2025}. This architecture balances local energy constraints with global Quality of Service (QoS).

\subsection{Healthcare and Wearables}
\label{sec:healthcare}
Medical applications face a dual constraint: extreme energy efficiency for implants and strict data privacy for wearables.

\subsubsection{Bio-signal Monitoring}
SNNs are increasingly applied to ECG classification and arrhythmia detection on FPGAs \cite{karamimanesh2025fpga}. A June 2024 review further corroborates this trajectory, noting that SNN-based implementations are rapidly closing the accuracy gap with DNNs while maintaining a fraction of the power budget \cite{ChoiHansol2024}. By processing raw bio-signals locally, these systems ensure that sensitive waveforms are never transmitted wirelessly, significantly reducing the attack surface for data breaches \cite{choi2024spiking, xing2022accurate}.

\subsubsection{Implantable Interfaces}
For intracortical neural decoding, heat dissipation is the limiting factor to prevent tissue damage. An April 2025 study on adaptive pruning of SNNs demonstrated \textbf{sub-$\mu$W power consumption} (0.18 $\mu$W/timestep) \cite{Rivelli2025}. This level of thermal efficiency is unattainable with standard Von Neumann architectures and represents a key enabling technology for long-term brain-machine interfaces.

\subsection{Industrial Detection and Smart Infrastructure}
\label{sec:industrial}
In large-scale infrastructure, the primary bottlenecks are bandwidth saturation and data sovereignty.

\subsubsection{Industrial Fault Diagnosis (IFD)}
Transmitting high-frequency vibration data from thousands of machines to a central cloud is often impractical due to bandwidth costs. SNNs enable \textbf{Edge Learning}, where models adapt to new defects locally. Recent methodologies demonstrate SNN model strengthening for surface defect detection directly on the edge node \cite{ZhouDeming2023, Alzarooni2025}. This eliminates the need to expose proprietary production line data to external networks.

\subsubsection{Smart City and Public Safety}
In traffic management, transmitting raw video streams saturates networks. SNNs allow for ``Semantic Compression,'' where only metadata (e.g., ``accident detected'') is transmitted rather than raw pixels \cite{DolatAbadi2025}. 
Beyond traffic, SNN-equipped UAVs are proving vital for \textbf{disaster evacuation monitoring} \cite{PalOsim2023}. In such scenarios, reliance on potentially damaged cellular infrastructure is risky, necessitating the autonomous, on-board intelligence that SNNs provide. Furthermore, by processing video at the sensor level and discarding personally identifiable information instantly, these systems inherently adhere to stricter privacy standards \cite{Wolniak2024, PuglieseViloria2024}.

\vspace{0.5em}
\noindent
\textit{Across these domains, a clear pattern emerges regarding the viability of SNNs. Spiking networks demonstrate distinct superiority in scenarios characterized by \textbf{event-driven dynamics, critical latency requirements, and severe power constraints} (e.g., neuromorphic control, always-on sensing). Conversely, in tasks requiring high-precision static pattern recognition with relaxed power budgets (e.g., static ImageNet classification), the advantages of SNNs over traditional CNNs remain less pronounced.}

\section{Grand Challenges: Systemic Friction}
\label{sec:challenges}

Despite the theoretical allure of Spiking Neural Networks (SNNs) for edge intelligence, their transition from academic curiosities to mainstream deployment is impeded by significant \textit{systemic friction}. This friction arises not merely from engineering bugs, but from a fundamental mismatch between current deep learning paradigms, silicon limitations, and the discrete nature of spiking dynamics. We categorize these challenges into three critical dimensions: the algorithmic learning dilemma, the hardware-software gap, and the operational realities of deployment. Table \ref{tab:challenges_thrusts} provides a structured taxonomy of these barriers, mapping specific technical pain points (e.g., dead neurons, memory wall) to current research thrusts and representative literature.

\begin{table*}[t]
    \centering
    \small 
    \caption{Major Challenges in Deploying SNNs at the Edge and Current Research Thrusts}
    \label{tab:challenges_thrusts}
    \renewcommand{\arraystretch}{1.3} 
    
    \begin{tabularx}{\linewidth}{@{} l >{\raggedright\hsize=0.8\hsize}X >{\raggedright\hsize=1.2\hsize}X c @{}}
    \toprule 
    \textbf{Challenge Area} & \textbf{Specific Challenge} & \textbf{Current Research Approaches \& Thrusts} & \textbf{Key Refs.} \\
    \midrule 
    
    \multirow{6}{*}{\textbf{\shortstack[l]{Training\\Complexity}}} 
        & Non-differentiable spike events 
        & \textbf{Surrogate Gradients (SG)}; Event-Driven Learning (STD-ED, MPD-ED); Differentiable Spike (Dspike) 
        & \cite{WeiWenjie2024, ZhouChenlin2024} \\
        
        & Gradient vanishing \& ``Dead neuron'' 
        & \textbf{Learnable SGs}; Adaptive Gradient Rules; Threshold-dependent Batch Norm (tdBN); Activity regularization 
        & \cite{JiangJiaqiang2025, ZhouChenlin2024} \\
        
        & Bio-plausible scalability (STDP) 
        & Hybrid learning (e.g., R-STDP); Supervised STDP variants; Combining STDP with backprop 
        & \cite{HuoBingqiang2025, tavanaei2019deep} \\
    \addlinespace[8pt] 
    
    \textbf{Scalability} 
        & Performance on large datasets (e.g., ImageNet) 
        & \textbf{ANN-to-SNN Conversion}; SNN-specific deep architectures (Spiking Transformers); Hybrid models 
        & \cite{HaoZecheng2024, YuKairong2025} \\
    \addlinespace[8pt]
    
    \multirow{3}{*}{\textbf{\shortstack[l]{Hardware\\Constraints}}} 
        & Memory Wall \& Fan-in/out limits 
        & \textbf{Co-design Frameworks} (e.g., QUEST); In-memory computing; SNN-specific pruning/quantization (OSBC) 
        & \cite{LiSai2025, ShiLianfeng2025} \\
        
        & Inter-chip comm. \& Heterogeneity 
        & Optimized NoC designs; Standardized IR (e.g., NIR); Platform-agnostic frameworks (Lava) 
        & \cite{Seekings2024, XiaoChao2022} \\
    \addlinespace[8pt]
    
    \textbf{Energy/Latency} 
        & Accuracy vs. Efficiency Trade-offs 
        & \textbf{Joint Optimization} (e.g., SNN4Agents); Edge-cloud collaboration (ECC-SNN); Early exit mechanisms 
        & \cite{PutraRachmad2024, YuDi2025} \\
    \addlinespace[8pt]
    
    \textbf{Standardization} 
        & Fragmented ecosystem \& Benchmarks 
        & \textbf{Unifying Frameworks} (Lava, Nengo); Federated SNN learning (FedLEC); Neuromorphic benchmarking suites 
        & \cite{Seekings2024, YuDongfang2024} \\
    
    \bottomrule
    \end{tabularx}
\end{table*}

\subsection{The Learning Dilemma: Optimization in a Discrete World}
The primary obstacle impeding SNN adoption is the algorithmic struggle to train deep, spiking architectures effectively.

\subsubsection{The Fundamental Schism of Non-Differentiability}
A fundamental schism exists between the gradient-based optimization dominant in Deep Learning (e.g., Backpropagation) and the discrete dynamics of SNNs. The generation of a spike is inherently an all-or-nothing event, typically modeled by a non-differentiable Heaviside step function. This discrete nature renders direct differentiation impossible, breaking the chain rule required for standard backpropagation.
While \textbf{Surrogate Gradients (SG)}—which approximate the spike derivative with a continuous function (e.g., sigmoid or arctan) during the backward pass—have become the standard workaround, they introduce a gradient mismatch. This approximation error accumulates as network depth increases, often leading to suboptimal convergence compared to ANNs.

\subsubsection{Training Stability: The ``Dead Neuron'' and BPTT}
Deep SNNs face severe stability issues. First, the \textbf{``Dead Neuron'' problem} implies that neurons may fail to cross the firing threshold due to poor initialization or insufficient stimulus, rendering them silent \cite{MaassWolfgang2015}. Unlike ReLUs in ANNs, a silent spiking neuron provides no gradient information and consumes time-steps without contributing to inference. 
Second, training relies on \textbf{Backpropagation Through Time (BPTT)} to capture temporal dependencies. BPTT unfolds the network over time steps, drastically increasing the memory footprint and exacerbating vanishing/exploding gradient problems, making the training of ultra-deep SNNs (e.g., ResNet-100+) computationally prohibitive.

\subsubsection{Scalability and the Conversion Debate}
A contentious debate persists regarding the optimal path to scalability:
\begin{itemize}
    \item \textbf{ANN-to-SNN Conversion:} This approach leverages mature ANN training pipelines and converts weights to SNNs. While stable, it often suffers from high latency (requiring long simulation windows to approximate rate coding) and fails to exploit the rich temporal dynamics of single spikes \cite{Schmolli2025}.
    \item \textbf{Direct Training:} Direct optimization (via SG) captures temporal features and enables low-latency inference. However, it struggles with the scalability issues mentioned above.
\end{itemize}
Consequently, while SNNs excel on neuromorphic benchmarks (e.g., CIFAR10-DVS), they struggle to match State-of-the-Art (SOTA) ANN accuracy on large-scale static datasets like ImageNet without incurring excessive latency costs.

\subsection{The Hardware-Software Gap: Breaking the Memory Wall}
The theoretical energy efficiency of SNNs is often predicated on idealized hardware. In reality, physical implementation faces strict constraints regarding memory hierarchy and interconnects.

\subsubsection{The State-Memory Wall: A Double Bottleneck}
SNNs introduce a unique challenge compared to stateless DNNs: the requirement to maintain the dynamic state (membrane potential, $V_{mem}$) of every neuron at every timestep \cite{Schuman2017}. This creates a ``double bottleneck'' of storing both synaptic weights and neuron states.

\begin{figure*}[t]
    \centering
    \includegraphics[width=0.98\linewidth]{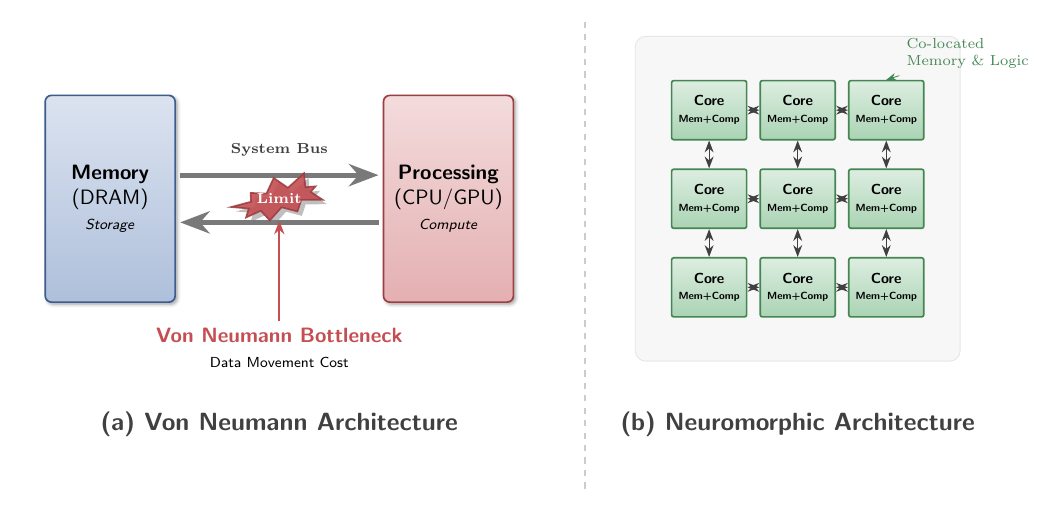}
    
    \caption{\textbf{Architectural Comparison.} (a) Traditional Von Neumann architecture separates memory and processing. The data bus (shown as parallel lines) becomes a bottleneck. (b) Neuromorphic architecture integrates memory and computation (M+C) into distributed cores.}
    \label{fig:von_neumann_vs_neuromorphic}
\end{figure*}

As visually explained in Fig.~\ref{fig:von_neumann_vs_neuromorphic}, solving this requires architectural divergence:
\begin{itemize}
    \item \textbf{Von Neumann Bottleneck (Fig.~\ref{fig:von_neumann_vs_neuromorphic}a):} Traditional architectures separate processing units (CPU/GPU) from memory. Data must traverse the system bus, leading to the ``Memory Wall,'' where energy consumption is dominated by data movement ($100\times \sim 1000\times$ higher than computation) rather than arithmetic logic.
    \item \textbf{Neuromorphic Locality (Fig.~\ref{fig:von_neumann_vs_neuromorphic}b):} Neuromorphic chips (e.g., Intel Loihi, TrueNorth) adopt a non-Von Neumann approach by co-locating memory and computation within distributed cores. By storing states locally (SRAM), they minimize off-chip data movement.
\end{itemize}

However, this locality introduces a capacity constraint. High-speed on-chip SRAM is expensive and limited \cite{davies2018loihi, Pei2019Tianjic, le2025review, hellenbrand2024progress, pan2024recent}. When deep SNN models exceed on-chip capacity, the system is forced to access off-chip DRAM, causing energy consumption to spike drastically and negating the benefits of event-driven processing.

\subsubsection{Connectivity and Fragmentation}
\textbf{Fan-in/Fan-out Sparsity:} Unlike the all-to-all connectivity in theoretical models, physical chips have limited routing resources. Mapping dense networks onto sparse hardware graphs necessitates ``multi-hop'' spike routing \cite{XueJianwei2023}, which introduces non-deterministic latency and congestion.

\textbf{The Software Stack Void:} Perhaps the most critical barrier is the absence of a unified software stack comparable to NVIDIA's CUDA. The landscape is fragmented into heterogeneous architectures (asynchronous, mixed-signal, digital), preventing the development of standardized compilation toolchains \cite{Vogginger2024}. This forces researchers into labor-intensive, hardware-specific optimization.

\subsection{The Deployment Reality: Security and Operational Trade-offs}
As SNNs migrate from labs to exposed edge environments, new operational vulnerabilities emerge.

\subsubsection{Adversarial Robustness: The Temporal Vulnerability}
SNNs were historically hypothesized to possess inherent robustness due to stochasticity and discrete filtering. However, recent scrutiny reveals a false sense of security. It has been demonstrated that SNNs are highly vulnerable to \textbf{temporal perturbations} \cite{Liang2021Exploring}. Attackers can imperceptibly shift the timing of input spikes, causing membrane potentials to miss firing thresholds. Recent surveys indicate that gradient-based attacks can be adapted to the temporal domain, necessitating robust defense mechanisms like discrete adversarial training \cite{baghbaderani2025adversarial, ghosh2025toward}.

\subsubsection{Privacy and Side-Channels}
While Federated Neuromorphic Learning (FNL) offers privacy by transmitting only weight updates \cite{skatchkovsky2020federated}, hardware implementations remain vulnerable. Physical side-channel attacks can monitor the distinct power signatures of spike events to reverse-engineer input stimuli or model architectures \cite{Huang2021SideChannel}, demanding hardware-level masking countermeasures.

\subsubsection{The Energy-Accuracy Trade-off}
Finally, deployment involves a harsh trade-off. Achieving the ultra-low energy promised by SNNs often requires aggressive quantization (e.g., 4-bit weights) or extremely short simulation windows, which can severely degrade accuracy. Frameworks like ECC-SNN attempt to mitigate this by offloading difficult samples to the cloud; however, this reintroduces a dependency on network connectivity.

\vspace{0.5em}
\noindent \textit{Ultimately, the future of SNNs does not lie in simply replacing DNNs for all tasks, but in identifying and dominating the \textbf{``Niche of Spatio-Temporal Sparsity''}—scenarios involving asynchronous event streams, ultra-low latency control loops, and constrained power budgets. The ``Grand Challenge'' is to bridge the schism between software flexibility and hardware primitives, transforming theoretical efficiency into operational reality.}

\section{Future Roadmap: Towards Ubiquitous Intelligence}
\label{sec:future}

As we look beyond the current hurdles, the trajectory of SNNs in edge computing is not merely about incremental improvements but represents a fundamental architectural shift. If the previous section outlined the friction points, this roadmap serves as a blueprint for the next generation of edge intelligence. We structure this vision into three cohesive pillars: the \textit{Algorithmic Frontier}, the \textit{Hardware-Software Continuum}, and the \textit{Convergence} of emerging technologies.

\begin{figure*}[t]
    \centering
    \includegraphics[width=0.85\linewidth]{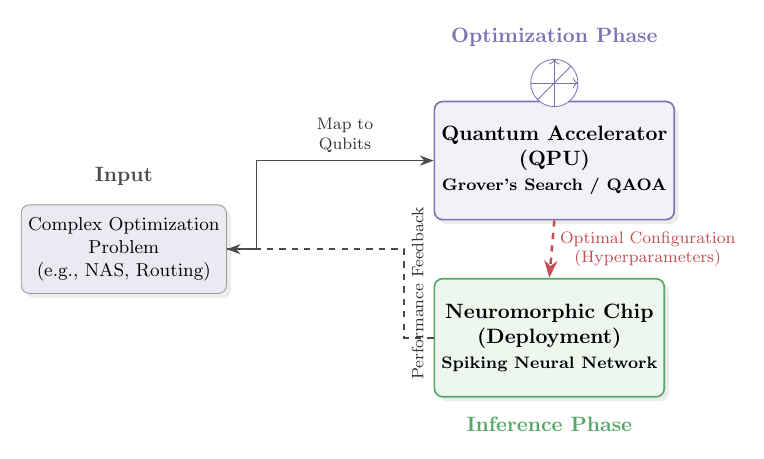}
    \caption{\textbf{Conceptual Quantum-Neuromorphic Synergy.} Quantum algorithms (e.g., QAOA) are utilized to solve NP-hard optimization problems inherent in SNN design, such as Neural Architecture Search (NAS) or synaptic routing, enabling the deployment of highly optimized SNNs on neuromorphic hardware.}
    \label{fig:quantum_synergy}
\end{figure*}

\subsection{The Algorithmic Frontier: From Rules to Reasoning}
To bridge the gap between biological efficiency and Artificial General Intelligence (AGI), we must move beyond simple classification tasks toward complex spatio-temporal reasoning \cite{Rathi2023}. This leap requires a symbiotic evolution of learning rules and model scale.

\subsubsection{Unifying Learning Paradigms}
The dichotomy between biological plausibility and gradient-based optimization must be resolved. Future algorithms will likely employ a hybrid strategy: using \textbf{Advanced Surrogate Gradients} for offline meta-learning of deep architectures (e.g., Spiking Transformers), while leveraging biologically inspired rules (such as advanced STDP variants) for rapid, energy-efficient \textbf{On-Chip Adaptation} \cite{PutraRachmad2025}. The goal is to develop ``Direct Training'' methods robust to the vanishing gradient problem, enabling SNNs to capture long-range dependencies without the excessive computational cost of BPTT \cite{YuKairong2025, Aydin2024}.

\subsubsection{Large-Scale Spiking Models}
Just as LLMs have revolutionized NLP, the era of \textbf{Large Spiking Models (LSMs)} is approaching. By integrating mechanisms like the Temporal Shift Module and attention mechanisms directly into the spiking domain, future models will orchestrate complex multimodal tasks on the edge. This shifts the paradigm from static datasets to dynamic, event-stream processing, where the sparsity of SNNs provides a decisive advantage in FLOPs reduction \cite{MaassWolfgang2015}.

\subsection{The Hardware-Software Continuum}
The potential of algorithms can only be realized if the underlying substrate—both physical and virtual—is optimized for event-driven dynamics.

\subsubsection{Next-Gen Neuromorphic Hardware}
The integration of \textbf{Emerging Non-Volatile Memories (eNVMs)} is pivotal. Devices such as \textbf{Memristors (RRAM)} and \textbf{Phase-Change Memory (PCM)} allow for the physical emulation of high-fan-out synaptic connectivity~\cite{syed2025phase, wang2025memristor, yu2018neuro}. Unlike traditional Von Neumann architectures, these devices enable analog \textit{Compute-in-Memory (CIM)}, where synaptic accumulation occurs directly within the memory array, virtually eliminating the energy cost of data movement \cite{Seekings2025}.

\subsubsection{The Neuromorphic OS: Orchestration and Virtualization}
Perhaps the most critical yet underexplored frontier is the system-level software stack. Current cloud orchestration frameworks (e.g., Kubernetes) suffer from a fundamental \textbf{``Sync-Async Mismatch''}: they force asynchronous, fine-grained spike streams into inefficient, synchronous fixed-time batches. This mismatch creates a \textbf{``Bottleneck of Causality,''} where the latency benefits of SNNs are negated by the waiting time of the orchestration layer. 

To address this, future edge systems must evolve into a true \textbf{Neuromorphic OS} underpinned by three core mechanisms:
\begin{enumerate}
    \item \textbf{Event-Driven Scheduling:} A transition from frame-based scheduling to interrupt-driven policies that respect the causality of neural events \cite{davies2018loihi, muller2022scalable}.
    \item \textbf{Dynamic Heterogeneity:} Schedulers must quantify the \textbf{\textit{Price of Converting} (PoC)}—the latency and energy cost of transcoding between spiking (neuromorphic) and non-spiking (CPU/GPU) domains. Decisions to offload tasks should be based on minimizing this PoC dynamically \cite{Pei2019Tianjic, wang2021end, Aydin2024}.
    \item \textbf{SLA-Aware Spiking Containerization:} To enable multi-tenancy, we envision lightweight virtualization technologies optimized for stateful neurons. This requires novel mechanisms for ``State Saving/Restoring'' of membrane potentials, allowing context switching between SNN models without performance degradation \cite{xue2024hardware, beloglazov2012energy, shrestha2022survey}.
\end{enumerate}

\subsection{The Convergence: Quantum, 6G, and Ethics}
SNNs will not exist in isolation but will serve as the cognitive engine within a broader technological ecosystem.

\subsubsection{The 6G Nexus}
As 6G networks introduce ultra-low latency and semantic communications, SNNs are positioned to become the native processing dialect of the network edge \cite{Zeydan2025, Barbieri2025}. Comprehensive 6G visions and surveys have highlighted neuromorphic computing as a key enabler for processing the unprecedented volume of sensor data directly at the infrastructure level \cite{WangChengXiang2023}.

\subsubsection{Quantum-Neuromorphic Synergy}
While currently exploratory, the intersection of quantum computing and SNNs offers a path to transcend classical limitations. As illustrated in Fig.~\ref{fig:quantum_synergy}, we propose a synergistic paradigm where Quantum Machine Learning (QML) algorithms (e.g., QAOA) serve as \textbf{offline accelerators}. These quantum agents solve NP-hard optimization problems inherent in SNN design—such as Neural Architecture Search (NAS) or complex synaptic routing. The optimized configuration is then mapped onto neuromorphic hardware for real-time, ultra-low-power inference \cite{Narottama2023}.

\subsubsection{Ethical and Secure Intelligence}
The ubiquity of ``always-on'' edge AI demands rigorous ethical guardrails. Beyond efficiency, SNNs must be designed for privacy (via Federated Learning, e.g., FedLEC \cite{YuDongfang2024}) and resilience against adversarial attacks targeting spike timing \cite{ZhouDeming2023, WangXubin2025}.

\bigskip
\noindent \textit{ Ultimately, we envision that the strict boundary between SNNs and DNNs will dissolve. The future of edge intelligence lies in a \textbf{``Unified Neuromorphic Continuum,''} where SNNs function as the ``ultra-low power gear'' for continuous monitoring, seamlessly shifting to the DNN ``high-performance gear'' only when deeper cognitive resolution is required. This seamless gear-shifting represents the holy grail of adaptive edge computing.}

\section{Conclusion}
\label{sec:conclusion}

This survey began by interrogating the ``Deployment Paradox'' facing neuromorphic computing: the tension between the theoretical efficiency of bio-inspired models and the practical rigidity of contemporary von Neumann hardware. Throughout this review, we have demonstrated that this paradox is not an inherent flaw, but a transitional friction that is effectively resolvable through the lens of hardware-software co-design. By reconciling algorithmic sparsity with physical silicon constraints, we have shown that Spiking Neural Networks (SNNs) are transitioning from theoretical curiosities to robust operational realities.

Crucially, the significance of SNNs extends beyond mere energy conservation. They represent a fundamental paradigm shift from ``processing data''—the legacy of static, frame-based analysis—to ``processing changes.'' This event-driven philosophy aligns computation with the dynamic nature of the physical world, offering a path to fundamentally eliminate the temporal redundancy that plagues conventional frame-based Deep Learning.

However, realizing this potential requires more than incremental optimization. As we argued, the immediate frontier for the community lies in dismantling the ``Sync-Async Mismatch.'' The development of a unified Neuromorphic OS is imperative to orchestrate asynchronous spiking workloads within the synchronous infrastructure of modern computing systems. Only by abstracting these hardware complexities can we enable the widespread adoption of neuromorphic solutions.

Ultimately, as we stand on the precipice of the 6G era and the Internet of Everything, the demand for pervasive, low-latency intelligence has never been greater. SNNs are poised to transcend their role as a niche technology to become the ``Green Cognitive Substrate'' of our digital infrastructure—ensuring that the future of edge intelligence is not only ubiquitous but also sustainable.

\bibliographystyle{IEEEtran}
\bibliography{ref}

@misc{BrainChipAkida,
  title={Akida Neural Processor IP Technical Reference Manual},
  author={{BrainChip Holdings Ltd}},
  year={2024},
  note={Available at: https://brainchip.com/akida-neural-processor-soc/}
}

@misc{IntelHalaPoint,
  title={Intel Builds World’s Largest Neuromorphic System to Enable More Sustainable AI},
  author={{Intel Labs}},
  year={2024},
  note={Press Release. Available at: https://www.intel.com/}
}

@article{botta2016integration,
  title={Integration of cloud computing and internet of things: a survey},
  author={Botta, Alessio and De Donato, Walter and Persico, Valerio and Pescap{\'e}, Antonio},
  journal={Future generation computer systems},
  volume={56},
  pages={684--700},
  year={2016},
  publisher={Elsevier}
}

@article{satyanarayanan2017emergence,
  title={The emergence of edge computing},
  author={Satyanarayanan, Mahadev},
  journal={Computer},
  volume={50},
  number={1},
  pages={30--39},
  year={2017},
  publisher={IEEE}
}

@article{chen2019deep,
  title={Deep learning with edge computing: A review},
  author={Chen, Jiasi and Ran, Xukan},
  journal={Proceedings of the IEEE},
  volume={107},
  number={8},
  pages={1655--1674},
  year={2019},
  publisher={IEEE}
}

@article{rydning2018digitization,
  title={The digitization of the world from edge to core},
  author={Rydning, David Reinsel-John Gantz-John and Reinsel, John and Gantz, John},
  journal={Framingham: International Data Corporation},
  volume={16},
  pages={1--28},
  year={2018}
}

@book{rabaey2002digital,
  title={Digital integrated circuits},
  author={Rabaey, Jan M and Chandrakasan, Anantha and Nikolic, Borivoje},
  volume={2},
  year={2002},
  publisher={Prentice Hall Englewood Cliffs}
}

@inproceedings{horowitz20141,
  title={1.1 computing's energy problem (and what we can do about it)},
  author={Horowitz, Mark},
  booktitle={2014 IEEE International Solid-State Circuits Conference Digest of Technical Papers (ISSCC)},
  pages={10--14},
  year={2014},
  organization={IEEE}
}

@inproceedings{shafique2014eda,
  title={The EDA challenges in the dark silicon era: Temperature, reliability, and variability perspectives},
  author={Shafique, Muhammad and Garg, Siddharth and Henkel, J{\"o}rg and Marculescu, Diana},
  booktitle={Proceedings of the 51st Annual Design Automation Conference},
  pages={1--6},
  year={2014}
}

@article{hardavellas2011toward,
  title={Toward dark silicon in servers},
  author={Hardavellas, Nikos and Ferdman, Michael and Falsafi, Babak and Ailamaki, Anastasia},
  journal={IEEE Micro},
  volume={31},
  number={4},
  pages={6--15},
  year={2011},
  publisher={IEEE}
}

@inproceedings{taylor2012dark,
  title={Is dark silicon useful? Harnessing the four horsemen of the coming dark silicon apocalypse},
  author={Taylor, Michael B},
  booktitle={Proceedings of the 49th annual design automation conference},
  pages={1131--1136},
  year={2012}
}

@inproceedings{esmaeilzadeh2011dark,
  title={Dark silicon and the end of multicore scaling},
  author={Esmaeilzadeh, Hadi and Blem, Emily and St. Amant, Renee and Sankaralingam, Karthikeyan and Burger, Doug},
  booktitle={Proceedings of the 38th annual international symposium on Computer architecture},
  pages={365--376},
  year={2011}
}

@article{wang2021end,
  title={End-to-end implementation of various hybrid neural networks on a cross-paradigm neuromorphic chip},
  author={Wang, Guanrui and Ma, Songchen and Wu, Yujie and Pei, Jing and Zhao, Rong and Shi, Luping},
  journal={Frontiers in neuroscience},
  volume={15},
  pages={615279},
  year={2021},
  publisher={Frontiers Media SA}
}

@article{shrestha2022survey,
  title={A survey on neuromorphic computing: Models and hardware},
  author={Shrestha, Amar and Fang, Haowen and Mei, Zaidao and Rider, Daniel Patrick and Wu, Qing and Qiu, Qinru},
  journal={IEEE Circuits and Systems Magazine},
  volume={22},
  number={2},
  pages={6--35},
  year={2022},
  publisher={IEEE}
}

@inproceedings{xue2024hardware,
  title={Hardware-assisted virtualization of neural processing units for cloud platforms},
  author={Xue, Yuqi and Liu, Yiqi and Nai, Lifeng and Huang, Jian},
  booktitle={2024 57th IEEE/ACM International Symposium on Microarchitecture (MICRO)},
  pages={1--16},
  year={2024},
  organization={IEEE}
}

@article{muller2022scalable,
  title={A scalable approach to modeling on accelerated neuromorphic hardware},
  author={M{\"u}ller, Eric and Arnold, Elias and Breitwieser, Oliver and Czierlinski, Milena and Emmel, Arne and Kaiser, Jakob and Mauch, Christian and Schmitt, Sebastian and Spilger, Philipp and Stock, Raphael and others},
  journal={Frontiers in neuroscience},
  volume={16},
  pages={884128},
  year={2022},
  publisher={Frontiers}
}

@article{beloglazov2012energy,
  title={Energy-aware resource allocation heuristics for efficient management of data centers for cloud computing},
  author={Beloglazov, Anton and Abawajy, Jemal and Buyya, Rajkumar},
  journal={Future generation computer systems},
  volume={28},
  number={5},
  pages={755--768},
  year={2012},
  publisher={Elsevier}
}

@article{tuli2019fogbus,
  title={Fogbus: A blockchain-based lightweight framework for edge and fog computing},
  author={Tuli, Shreshth and Mahmud, Redowan and Tuli, Shikhar and Buyya, Rajkumar},
  journal={Journal of Systems and Software},
  volume={154},
  pages={22--36},
  year={2019},
  publisher={Elsevier}
}

@article{gupta2017ifogsim,
  title={iFogSim: A toolkit for modeling and simulation of resource management techniques in the Internet of Things, Edge and Fog computing environments},
  author={Gupta, Harshit and Vahid Dastjerdi, Amir and Ghosh, Soumya K and Buyya, Rajkumar},
  journal={Software: Practice and Experience},
  volume={47},
  number={9},
  pages={1275--1296},
  year={2017},
  publisher={Wiley Online Library}
}

@article{shi2016edge,
  title={Edge computing: Vision and challenges},
  author={Shi, Weisong and Cao, Jie and Zhang, Quan and Li, Youhuizi and Xu, Lanyu},
  journal={IEEE Internet of Things Journal},
  volume={3},
  number={5},
  pages={637--646},
  year={2016},
  publisher={IEEE}
}

@article{zhou2019edge,
  title={Edge intelligence: Paving the last mile of artificial intelligence with edge computing},
  author={Zhou, Zhi and Chen, Xu and Li, En and Zeng, Liekang and Luo, Ke and Zhang, Junshan},
  journal={Proceedings of the IEEE},
  volume={107},
  number={8},
  pages={1738--1762},
  year={2019},
  publisher={IEEE}
}

@article{buyya2018manifesto,
  title={A manifesto for future generation cloud computing: Research directions for the next decade},
  author={Buyya, Rajkumar and Srirama, Satish Narayana and Casale, Giuliano and Calheiros, Rodrigo and Simmhan, Yogesh and Varghese, Blesson and Gelenbe, Erol and Javadi, Bahman and Vaquero, Luis Miguel and Netto, Marco AS and others},
  journal={ACM computing surveys (CSUR)},
  volume={51},
  number={5},
  pages={1--38},
  year={2018},
  publisher={ACM New York, NY, USA}
}

@article{yu2018neuro,
  title={Neuro-inspired computing with emerging nonvolatile memorys},
  author={Yu, Shimeng},
  journal={Proceedings of the IEEE},
  volume={106},
  number={2},
  pages={260--285},
  year={2018},
  publisher={IEEE}
}

@article{wang2025memristor,
  title={Memristor-Based Spiking Neuromorphic Systems Toward Brain-Inspired Perception and Computing},
  author={Wang, Xiangjing and Zhu, Yixin and Zhou, Zili and Chen, Xin and Jia, Xiaojun},
  journal={Nanomaterials},
  volume={15},
  number={14},
  pages={1130},
  year={2025},
  publisher={MDPI}
}

@article{syed2025phase,
  title={Phase-Change Memory for In-Memory Computing},
  author={Syed, Ghazi Sarwat and Le Gallo, Manuel and Sebastian, Abu},
  journal={Chemical Reviews},
  year={2025},
  publisher={ACS Publications}
}

@INPROCEEDINGS{ghosh2025toward,
  author={Ghosh, Archisman and Ghosh, Swaroop},
  booktitle={2025 IEEE/ACM International Conference On Computer-Aided Design (ICCAD)}, 
  title={Invited Paper: Toward Secure In-Sensor Intelligence: Threats and Defenses in SNNs}, 
  year={2025},
  volume={},
  number={},
  pages={1-7},
  doi={10.1109/ICCAD66269.2025.11240674}}

@inproceedings{baghbaderani2025adversarial,
  title={Adversarial Attacks and Defense Mechanisms in Spiking Neural Networks: A Comprehensive Review},
  author={Baghbaderani, Bahareh Kaviani and Shahinzadeh, Hossein},
  booktitle={2025 29th International Computer Conference, Computer Society of Iran (CSICC)},
  pages={01--07},
  year={2025},
  organization={IEEE}
}

@article{pan2024recent,
  title={Recent progress of non-volatile memory devices based on two-dimensional materials},
  author={Pan, Jiong and Wang, Zeda and Zhao, Bingchen and Yin, Jiaju and Guo, Pengwen and Yang, Yi and Ren, Tian-Ling},
  journal={Chips},
  volume={3},
  number={4},
  pages={271--295},
  year={2024},
  publisher={MDPI}
}

@article{hellenbrand2024progress,
  title={Progress of emerging non-volatile memory technologies in industry},
  author={Hellenbrand, Markus and Teck, Isabella and MacManus-Driscoll, Judith L},
  journal={MRS communications},
  volume={14},
  number={6},
  pages={1099--1112},
  year={2024},
  publisher={Springer}
}

@inproceedings{le2025review,
  title={A review of memory wall for neuromorphic computing},
  author={Le, Dexter and Arig, Baran and Isik, Murat and Dikmen, I Can and Karadag, Teoman},
  booktitle={2025 IEEE 4th International Conference on Computing and Machine Intelligence (ICMI)},
  pages={1--6},
  year={2025},
  organization={IEEE}
}

@article{xing2022accurate,
  title={Accurate ECG classification based on spiking neural network and attentional mechanism for real-time implementation on personal portable devices},
  author={Xing, Yuxuan and Zhang, Lei and Hou, Zhixian and Li, Xiaoran and Shi, Yueting and Yuan, Yiyang and Zhang, Feng and Liang, Sen and Li, Zhenzhong and Yan, Liang},
  journal={Electronics},
  volume={11},
  number={12},
  pages={1889},
  year={2022},
  publisher={MDPI}
}

@article{choi2024spiking,
  title={Spiking neural networks for biomedical signal analysis},
  author={Choi, Sang Ho},
  journal={Biomedical Engineering Letters},
  volume={14},
  number={5},
  pages={955--966},
  year={2024},
  publisher={Springer}
}

@article{bing2018survey,
  title={A survey of robotics control based on learning-inspired spiking neural networks},
  author={Bing, Zhenshan and Meschede, Claus and R{\"o}hrbein, Florian and Huang, Kai and Knoll, Alois C},
  journal={Frontiers in neurorobotics},
  volume={12},
  pages={35},
  year={2018},
  publisher={Frontiers Media SA}
}

@article{mompo2024brain,
  title={Brain-inspired biomimetic robot control: a review},
  author={Momp{\'o} Alepuz, Adri{\`a} and Papageorgiou, Dimitrios and Tolu, Silvia},
  journal={Frontiers in Neurorobotics},
  volume={18},
  pages={1395617},
  year={2024},
  publisher={Frontiers Media SA}
}

@article{sudevan2025underwater,
  title={Underwater Image Enhancement by Convolutional Spiking Neural Networks},
  author={Sudevan, Vidya and Zayer, Fakhreddine and Kausar, Rizwana and Javed, Sajid and Karki, Hamad and De Masi, Giulia and Dias, Jorge},
  journal={arXiv preprint arXiv:2503.20485},
  year={2025}
}

@article{cheng2025comprehensive,
  title={A comprehensive multimodal benchmark of neuromorphic training frameworks for spiking neural networks},
  author={Cheng, Ying-Chao and Hu, Wang-Xin and He, Yu-Lin and Huang, Joshua Zhexue},
  journal={Engineering Applications of Artificial Intelligence},
  volume={159},
  pages={111543},
  year={2025},
  publisher={Elsevier}
}

@article{yarga2024neuromorphic,
  title={Neuromorphic Keyword Spotting with Pulse Density Modulation MEMS Microphones},
  author={Yarga, Sidi Yaya Arnaud and Wood, Sean UN},
  journal={arXiv preprint arXiv:2408.05156},
  year={2024}
}

@inproceedings{blouw2019benchmarking,
  title={Benchmarking keyword spotting efficiency on neuromorphic hardware},
  author={Blouw, Peter and Choo, Xuan and Hunsberger, Eric and Eliasmith, Chris},
  booktitle={Proceedings of the 7th annual neuro-inspired computational elements workshop},
  pages={1--8},
  year={2019}
}

@inproceedings{shi2024towards,
  title={Towards energy efficient spiking neural networks: An unstructured pruning framework},
  author={Shi, Xinyu and Ding, Jianhao and Hao, Zecheng and Yu, Zhaofei},
  booktitle={The Twelfth International Conference on Learning Representations},
  year={2024}
}

@article{hu2024bitsnns,
  title={BitSNNs: revisiting energy-efficient spiking neural networks},
  author={Hu, Yangfan and Zheng, Qian and Pan, Gang},
  journal={IEEE Transactions on Cognitive and Developmental Systems},
  volume={16},
  number={5},
  pages={1736--1747},
  year={2024},
  publisher={IEEE}
}

@inproceedings{nguyen2025sparsified,
  title={Sparsified Federated Learning With Spiking Neural Networks: Resistance Against Byzantine Attacks While Lowering Communication Traffics},
  author={Nguyen, Manh V and Zhao, Liang and Deng, Bobin and Wu, Shaoen},
  booktitle={2025 IEEE 101st Vehicular Technology Conference (VTC2025-Spring)},
  pages={1--5},
  year={2025},
  organization={IEEE}
}

@article{aksu2025privacy,
  title={Privacy in Federated Learning with Spiking Neural Networks},
  author={Aksu, Dogukan and del Rincon, Jesus Martinez and Alouani, Ihsen},
  journal={arXiv preprint arXiv:2511.21181},
  year={2025}
}

@inproceedings{abbasihafshejani2025spiking,
  title={Spiking Neural Networks in Vertical Federated Learning: Performance Trade-offs},
  author={Abbasihafshejani, Maryam and Maiti, Anindya and Jadliwala, Murtuza},
  booktitle={NOMS 2025-2025 IEEE Network Operations and Management Symposium},
  pages={1--5},
  year={2025},
  organization={IEEE}
}

@article{al2024survey,
  title={A survey on neuromorphic architectures for running artificial intelligence algorithms},
  author={Al Abdul Wahid, Seham and Asad, Arghavan and Mohammadi, Farah},
  journal={Electronics},
  volume={13},
  number={15},
  pages={2963},
  year={2024},
  publisher={MDPI}
}

@inproceedings{manna2023frameworks,
  title={Frameworks for SNNs: a review of data science-oriented software and an expansion of spyketorch},
  author={Manna, Davide L and Vicente-Sola, Alex and Kirkland, Paul and Bihl, Trevor J and Di Caterina, Gaetano},
  booktitle={International Conference on Engineering Applications of Neural Networks},
  pages={227--238},
  year={2023},
  organization={Springer}
}

@article{kulkarni2021benchmarking,
  title={Benchmarking the performance of neuromorphic and spiking neural network simulators},
  author={Kulkarni, Shruti R and Parsa, Maryam and Mitchell, J Parker and Schuman, Catherine D},
  journal={Neurocomputing},
  volume={447},
  pages={145--160},
  year={2021},
  publisher={Elsevier}
}

@article{akopyan2015truenorth,
  title={Truenorth: Design and tool flow of a 65 mw 1 million neuron programmable neurosynaptic chip},
  author={Akopyan, Filipp and Sawada, Jun and Cassidy, Andrew and Alvarez-Icaza, Rodrigo and Arthur, John and Merolla, Paul and Imam, Nabil and Nakamura, Yutaka and Datta, Pallab and Nam, Gi-Joon and others},
  journal={IEEE transactions on computer-aided design of integrated circuits and systems},
  volume={34},
  number={10},
  pages={1537--1557},
  year={2015},
  publisher={IEEE}
}

@article{balaji2019mapping,
  title={Mapping spiking neural networks to neuromorphic hardware},
  author={Balaji, Adarsha and Das, Anup and Wu, Yuefeng and Huynh, Khanh and Dell’Anna, Francesco G and Indiveri, Giacomo and Krichmar, Jeffrey L and Dutt, Nikil D and Schaafsma, Siebren and Catthoor, Francky},
  journal={IEEE Transactions on Very Large Scale Integration (VLSI) Systems},
  volume={28},
  number={1},
  pages={76--86},
  year={2019},
  publisher={IEEE}
}

@article{urgese2016optimizing,
  title={Optimizing network traffic for spiking neural network simulations on densely interconnected many-core neuromorphic platforms},
  author={Urgese, Gianvito and Barchi, Francesco and Macii, Enrico and Acquaviva, Andrea},
  journal={IEEE Transactions on Emerging Topics in Computing},
  volume={6},
  number={3},
  pages={317--329},
  year={2016},
  publisher={IEEE}
}

@ARTICLE{chen2022asurvey,
  author={Chen, Lu and Xiong, Xingzhong and Liu, Jun},
  journal={IEEE Access}, 
  title={A Survey of Intelligent Chip Design Research Based on Spiking Neural Networks}, 
  year={2022},
  volume={10},
  number={},
  pages={89663-89686},
  doi={10.1109/ACCESS.2022.3200454}}

@article{pehle2022brainscales,
  title={The BrainScaleS-2 accelerated neuromorphic system with hybrid plasticity},
  author={Pehle, Christian and Billaudelle, Sebastian and Cramer, Benjamin and Kaiser, Jakob and Schreiber, Korbinian and Stradmann, Yannik and Weis, Johannes and Leibfried, Aron and M{\"u}ller, Eric and Schemmel, Johannes},
  journal={Frontiers in Neuroscience},
  volume={16},
  pages={795876},
  year={2022},
  publisher={Frontiers Media SA}
}

@article{christensen20222022,
  title={2022 roadmap on neuromorphic computing and engineering},
  author={Christensen, Dennis V and Dittmann, Regina and Linares-Barranco, Bernabe and Sebastian, Abu and Le Gallo, Manuel and Redaelli, Andrea and Slesazeck, Stefan and Mikolajick, Thomas and Spiga, Sabina and Menzel, Stephan and others},
  journal={Neuromorphic Computing and Engineering},
  volume={2},
  number={2},
  pages={022501},
  year={2022},
  publisher={IOP Publishing}
}

@article{zhao2022spiking,
  title={Spiking CapsNet: A Spiking Neural Network with a Biologically Plausible Routing Rule between Capsules},
  author={Zhao, Dongcheng and Li, Yang and Zeng, Yi and Wang, Jihang and Zhang, Qian},
  journal={Information Sciences},
  volume={610},
  pages={1--13},
  year={2022},
  publisher={Elsevier}
}

@article{yao2020fully,
  title={Fully hardware-implemented memristor convolutional neural network},
  author={Yao, Peng and Wu, Huaqiang and Gao, Bin and Tang, Jianshi and Zhang, Qingtian and Zhang, Wenqiang and Yang, J Joshua and Qian, He},
  journal={Nature},
  volume={577},
  number={7792},
  pages={641--646},
  year={2020},
  publisher={Nature Publishing Group UK London}
}

@incollection{mayr2019spinnaker,
  title={SpiNNaker 2: A 10 million core processor system for brain simulation and machine learning-Keynote presentation},
  author={Mayr, Christian and Hoeppner, Sebastian and Furber, Steve},
  booktitle={Communicating Process Architectures 2017 \& 2018},
  pages={277--280},
  year={2019},
  publisher={IOS Press}
}

@inproceedings{orchard2021efficient,
  title={Efficient neuromorphic signal processing with Loihi 2},
  author={Orchard, Garrick and Frady, E Paxon and Rubin, Daniel Ben Dayan and Sanborn, Sophia and Shrestha, Sumit Bam and Sommer, Friedrich T and Davies, Mike},
  booktitle={2021 IEEE Workshop on Signal Processing Systems (SiPS)},
  pages={254--259},
  year={2021},
  organization={IEEE}
}

@inproceedings{song2024one,
  title={One-step spiking transformer with a linear complexity},
  author={Song, Xiaotian and Song, Andy and Xiao, Rong and Sun, Yanan},
  booktitle={Proceedings of the Thirty-Third International Joint Conference on Artificial Intelligence},
  pages={3142--3150},
  year={2024}
}

@inproceedings{bu2025inference,
  title={Inference-Scale Complexity in ANN-SNN Conversion for High-Performance and Low-Power Applications},
  author={Bu, Tong and Li, Maohua and Yu, Zhaofei},
  booktitle={Proceedings of the Computer Vision and Pattern Recognition Conference},
  pages={24387--24397},
  year={2025}
}

@article{ding2021optimal,
  title={Optimal ANN-SNN conversion for fast and accurate inference in deep spiking neural networks},
  author={Ding, Jianhao and Yu, Zhaofei and Tian, Yonghong and Huang, Tiejun},
  journal={arXiv preprint arXiv:2105.11654},
  year={2021}
}

@inproceedings{Shi2025,
  title={Spiking Brain Compression: Exploring One-Shot Post-Training Pruning and Quantization for Spiking Neural Networks},
  author={Shi, Lianfeng and Li, Ao and Ward-Cherrier, Benjamin},
  booktitle={OPT2025: 17th Annual Workshop on Optimization for Machine Learning},
  year={2025},
  note={arXiv:2506.03996}
}

@inproceedings{Amir2017,
  title={A Low Power, Fully Event-Based Gesture Recognition System},
  author={Amir, Arnon and Taba, Brian and Berg, David and Melano, Timothy and McKinstry, Jeffrey and Di Nolfo, Carmelo and Nayak, Tapan and Andreopoulos, Alexander and Garreau, Guillaume and Mendoza, Marcela and others},
  booktitle={Proceedings of the IEEE Conference on Computer Vision and Pattern Recognition (CVPR)},
  pages={7243--7252},
  year={2017}
}

@article{modha2023neural,
  title={Neural Inference at the Frontier of Energy, Space, and Time},
  author={Modha, Dharmendra S. and Akopyan, Filipp and Andreopoulos, Alexander and Appuswamy, Rathinakumar and Arthur, John V. and Cassidy, Andrew S. and Datta, Pallab and others},
  journal={Science},
  volume={382},
  number={6668},
  pages={329--335},
  year={2023},
  publisher={American Association for the Advancement of Science},
  doi={10.1126/science.adh1174}
}

@article{gallego2022event,
  title={Event-Based Vision: A Survey},
  author={Gallego, Guillermo and Delbr{\"u}ck, Tobi and Orchard, Garrick and Bartolozzi, Chiara and Taba, Brian and Censi, Andrea and Leutenegger, Stefan and others},
  journal={IEEE Transactions on Pattern Analysis and Machine Intelligence},
  volume={44},
  number={1},
  pages={154--180},
  year={2022},
  publisher={IEEE},
  doi={10.1109/TPAMI.2020.3008413}
}

@ARTICLE{rathi2023diet,
  author={Rathi, Nitin and Roy, Kaushik},
  journal={IEEE Transactions on Neural Networks and Learning Systems}, 
  title={DIET-SNN: A Low-Latency Spiking Neural Network With Direct Input Encoding and Leakage and Threshold Optimization}, 
  year={2023},
  volume={34},
  number={6},
  pages={3174-3182},
  doi={10.1109/TNNLS.2021.3111897}}

@article{davies2021advancing,
  title={Advancing Neuromorphic Computing with {Loihi}: A Survey of Results and Outlook},
  author={Davies, Mike and Wild, Andreas and Orchard, Garrick and Sandamirskaya, Yulia and Guerra, Gabriel A. Fonseca and Joshi, Prasad and Plank, Philipp and Risbud, Sumedh R.},
  journal={Proceedings of the IEEE},
  volume={109},
  number={5},
  pages={911--934},
  year={2021},
  publisher={IEEE},
  doi={10.1109/JPROC.2021.3067593}
}

@article{gygax2025elucidating,
  title={Elucidating the Theoretical Underpinnings of Surrogate Gradient Learning in Spiking Neural Networks},
  author={Gygax, Julia and Zenke, Friedemann},
  journal={Neural Computation},
  volume={37},
  number={5},
  pages={886--925},
  year={2025},
  publisher={MIT Press},
  doi={10.1162/neco_a_01732}
}

@phdthesis{frenkel2020bottom,
  title={Bottom-up and Top-down Neuromorphic Processor Design: Unveiling Roads to Embedded Cognition},
  author={Frenkel, Charlotte},
  school={Universit{\'e} catholique de Louvain},
  year={2020},
  address={Louvain-la-Neuve, Belgium}
}

@article{jo2010nanoscale,
  title={Nanoscale Memristor Device as Synapse in Neuromorphic Systems},
  author={Jo, Sung Hyun and Chang, Ting and Ebong, Idongesit and Bhadviya, Bhavitavya B. and Mazumder, Pinaki and Lu, Wei},
  journal={Nano Letters},
  volume={10},
  number={4},
  pages={1297--1301},
  year={2010},
  publisher={ACS Publications},
  doi={10.1021/nl904092h}
}

@article{guo2021neural,
  title={Neural Coding in Spiking Neural Networks: A Comparative Study for Robust Neuromorphic Systems},
  author={Guo, Wenzhe and Fouda, Mohammed E. and Eltawil, Ahmed M. and Salama, Khaled Nabil},
  journal={Frontiers in Neuroscience},
  volume={15},
  pages={638474},
  year={2021},
  publisher={Frontiers},
  doi={10.3389/fnins.2021.638474}
}

@article{auge2021survey,
  title={A survey of encoding techniques for signal processing in spiking neural networks},
  author={Auge, Daniel and Hille, Julian and Mueller, Etienne and Knoll, Alois},
  journal={Neural Processing Letters},
  volume={53},
  number={6},
  pages={4693--4710},
  year={2021},
  publisher={Springer}
}

@article{mostafa2017supervised,
  title={Supervised learning based on temporal coding in spiking neural networks},
  author={Mostafa, Hesham},
  journal={IEEE Transactions on Neural Networks and Learning Systems},
  volume={29},
  number={7},
  pages={3227--3235},
  year={2018},
  publisher={IEEE}
}

@article{gautrais1998rate,
  title={Rate coding versus temporal order coding: a theoretical approach},
  author={Gautrais, Jacques and Thorpe, Simon},
  journal={BioSystems},
  volume={48},
  number={1-3},
  pages={57--65},
  year={1998},
  publisher={Elsevier}
}

@article{thorpe2001spike,
  title={Spike-based strategies for rapid processing},
  author={Thorpe, Simon and Delorme, Arnaud and Van Rullen, Rufin},
  journal={Neural Networks},
  volume={14},
  number={6-7},
  pages={715--725},
  year={2001},
  publisher={Elsevier}
}

@article{brette2007simulation,
  title={Simulation of networks of spiking neurons: A review of tools and strategies},
  author={Brette, Romain and Rudolph, Michelle and Carnevale, Ted and Hines, Michael and Beeman, David and Bower, James M and Diesmann, Markus and Morrison, Abigail and Goodman, Philip H and Harris, Frederick C and others},
  journal={Journal of Computational Neuroscience},
  volume={23},
  number={3},
  pages={349--398},
  year={2007},
  publisher={Springer},
  note={Discusses the numerical stiffness and high computational cost of conductance-based models}
}

@article{furber2014spinnaker,
  title={The {SpiNNaker} Project},
  author={Furber, Steve B. and Galluppi, Francesco and Temple, Steve and Plana, Luis A.},
  journal={Proceedings of the IEEE},
  volume={102},
  number={5},
  pages={652--665},
  year={2014},
  publisher={IEEE},
  doi={10.1109/JPROC.2014.2304638}
}

@article{davies2018loihi,
  title={{Loihi}: A Neuromorphic Manycore Processor with On-Chip Learning},
  author={Davies, Mike and Srinivasa, Narayan and Lin, Tsung-Han and Chinya, Gautham and Cao, Yongqiang and Choday, Sri Harsha and Dimou, Georgios and others},
  journal={IEEE Micro},
  volume={38},
  number={1},
  pages={82--99},
  year={2018},
  publisher={IEEE},
  doi={10.1109/MM.2018.112130359}
}

@article{merolla2014million,
  title={A Million Spiking-Neuron Integrated Circuit with a Scalable Communication Network and Interface},
  author={Merolla, Paul A. and Arthur, John V. and Alvarez-Icaza, Rodrigo and Cassidy, Andrew S. and Sawada, Jun and Akopyan, Filipp and Jackson, Bryan L. and others},
  journal={Science},
  volume={345},
  number={6197},
  pages={668--673},
  year={2014},
  publisher={American Association for the Advancement of Science},
  doi={10.1126/science.1254642}
}

@article{zenke2018superspike,
  title={{SuperSpike}: Supervised Learning in Multilayer Spiking Neural Networks},
  author={Zenke, Friedemann and Ganguli, Surya},
  journal={Neural Computation},
  volume={30},
  number={6},
  pages={1514--1541},
  year={2018},
  publisher={MIT Press},
  doi={10.1162/neco_a_01086}
}

@inproceedings{shrestha2018slayer,
  title={{SLAYER}: Spike Layer Error Reassignment in Time},
  author={Shrestha, Sumit B. and Orchard, Garrick},
  booktitle={Advances in Neural Information Processing Systems},
  volume={31},
  pages={1412--1421},
  year={2018},
  publisher={Curran Associates, Inc.}
}

@article{bellec2020solution,
  title={A Solution to the Learning Dilemma for Recurrent Networks of Spiking Neurons},
  author={Bellec, Guillaume and Scherr, Franz and Subramoney, Anand and Hajek, Elias and Salaj, Darjan and Legenstein, Robert and Maass, Wolfgang},
  journal={Nature Communications},
  volume={11},
  number={1},
  pages={3625},
  year={2020},
  publisher={Nature Publishing Group},
  doi={10.1038/s41467-020-17236-y}
}

@inproceedings{li2021differentiable,
  title={Differentiable Spike: Rethinking Gradient-Descent for Training Spiking Neural Networks},
  author={Li, Yuhang and Guo, Yufei and Zhang, Shanghang and Deng, Shikuang and Hai, Yongqing and Gu, Shi},
  booktitle={Advances in Neural Information Processing Systems},
  volume={34},
  pages={23426--23439},
  year={2021},
  publisher={Curran Associates, Inc.}
}

@article{roy2019towards,
  title={Towards Spike-Based Machine Intelligence with Neuromorphic Computing},
  author={Roy, Kaushik and Jaiswal, Akhilesh and Panda, Priyadarshini},
  journal={Nature},
  volume={575},
  number={7784},
  pages={607--617},
  year={2019},
  publisher={Nature Publishing Group},
  doi={10.1038/s41586-019-1677-2}
}

@article{tavanaei2019deep,
  title={Deep Learning in Spiking Neural Networks},
  author={Tavanaei, Amirhossein and Ghodrati, Masoud and Kheradpisheh, Saeed Reza and Masquelier, Timoth{\'e}e and Maida, Anthony},
  journal={Neural Networks},
  volume={111},
  pages={47--63},
  year={2019},
  publisher={Elsevier},
  doi={10.1016/j.neunet.2018.12.002}
}

@article{richards2019deep,
  title={A Deep Learning Framework for Neuroscience},
  author={Richards, Blake A. and Lillicrap, Timothy P. and Beaudoin, Philippe and Bengio, Yoshua and Bogacz, Rafal and Christensen, Amelia and Clopath, Claudia and others},
  journal={Nature Neuroscience},
  volume={22},
  number={11},
  pages={1761--1770},
  year={2019},
  publisher={Nature Publishing Group},
  doi={10.1038/s41593-019-0520-2}
}

@article{neftci2019surrogate,
  title={Surrogate Gradient Learning in Spiking Neural Networks: Bringing the Power of Gradient-Based Optimization to Spiking Neural Networks},
  author={Neftci, Emre O. and Mostafa, Hesham and Zenke, Friedemann},
  journal={IEEE Signal Processing Magazine},
  volume={36},
  number={6},
  pages={51--63},
  year={2019},
  publisher={IEEE},
  doi={10.1109/MSP.2019.2931595}
}

@article{bi1998synaptic,
  title={Synaptic Modifications in Cultured Hippocampal Neurons: Dependence on Spike Timing, Synaptic Strength, and Postsynaptic Cell Type},
  author={Bi, Guo-qiang and Poo, Mu-ming},
  journal={Journal of Neuroscience},
  volume={18},
  number={24},
  pages={10464--10472},
  year={1998},
  publisher={Society for Neuroscience},
  doi={10.1523/JNEUROSCI.18-24-10464.1998}
}

@article{song2000competitive,
  title={Competitive Hebbian Learning through Spike-Timing-Dependent Synaptic Plasticity},
  author={Song, Sen and Miller, Kenneth D. and Abbott, Larry F.},
  journal={Nature Neuroscience},
  volume={3},
  number={9},
  pages={919--926},
  year={2000},
  publisher={Nature Publishing Group},
  doi={10.1038/78829}
}

@article{markram1997regulation,
  title={Regulation of Synaptic Efficacy by Coincidence of Postsynaptic {APs} and {EPSPs}},
  author={Markram, Henry and L{\"u}bke, Joachim and Frotscher, Michael and Sakmann, Bert},
  journal={Science},
  volume={275},
  number={5297},
  pages={213--215},
  year={1997},
  publisher={American Association for the Advancement of Science},
  doi={10.1126/science.275.5297.213}
}

@article{indiveri2011neuromorphic,
  title={Neuromorphic Silicon Neuron Circuits},
  author={Indiveri, Giacomo and Linares-Barranco, Bernab{\'e} and Hamilton, Tara Julia and van Schaik, Andr{\'e} and Etienne-Cummings, Ralph and Delbruck, Tobi and Liu, Shih-Chii and others},
  journal={Frontiers in Neuroscience},
  volume={5},
  pages={73},
  year={2011},
  publisher={Frontiers},
  doi={10.3389/fnins.2011.00073}
}

@article{mead1990neuromorphic,
  title={Neuromorphic Electronic Systems},
  author={Mead, Carver},
  journal={Proceedings of the IEEE},
  volume={78},
  number={10},
  pages={1629--1636},
  year={1990},
  publisher={IEEE},
  doi={10.1109/5.58356}
}

@book{mead1989analog,
  title={Analog {VLSI} Implementation of Neural Systems},
  editor={Mead, Carver and Ismail, Mohammed},
  series={The Kluwer International Series in Engineering and Computer Science},
  volume={80},
  year={1989},
  publisher={Springer},
  address={Boston, MA},
  doi={10.1007/978-1-4613-1639-8}
}

@article{brette2005adaptive,
  title={Adaptive Exponential Integrate-and-Fire Model as an Effective Description of Neuronal Activity},
  author={Brette, Romain and Gerstner, Wulfram},
  journal={Journal of Neurophysiology},
  volume={94},
  number={5},
  pages={3637--3642},
  year={2005},
  publisher={American Physiological Society},
  doi={10.1152/jn.00686.2005}
}

@article{naud2008firing,
  title={Firing Patterns in the Adaptive Exponential Integrate-and-Fire Model},
  author={Naud, Richard and Marcille, Nicolas and Clopath, Claudia and Gerstner, Wulfram},
  journal={Biological Cybernetics},
  volume={99},
  number={4-5},
  pages={335--347},
  year={2008},
  publisher={Springer},
  doi={10.1007/s00422-008-0264-7}
}

@article{jolivet2008benchmark,
  title={A Benchmark Test for a Quantitative Assessment of Simple Neuron Models},
  author={Jolivet, Renaud and Kobayashi, Ryota and Rauch, Alexander and Naud, Richard and Shinomoto, Shigeru and Gerstner, Wulfram},
  journal={Journal of Neuroscience Methods},
  volume={169},
  number={2},
  pages={417--424},
  year={2008},
  publisher={Elsevier},
  doi={10.1016/j.jneumeth.2007.11.006}
}

@article{lobo2020spiking,
  title={Spiking Neural Networks and Online Learning: An Overview and Perspectives},
  author={Lobo, Jesus L. and Del Ser, Javier and Bifet, Albert and Kasabov, Nikola},
  journal={Neural Networks},
  volume={121},
  pages={88--100},
  year={2020},
  publisher={Elsevier},
  doi={10.1016/j.neunet.2019.09.006}
}

@inproceedings{skatchkovsky2020federated,
  title={Federated Neuromorphic Learning of Spiking Neural Networks for Low-Power Edge Intelligence},
  author={Skatchkovsky, Nicolas and Jang, Hyeryung and Simeone, Osvaldo},
  booktitle={ICASSP 2020 - 2020 IEEE International Conference on Acoustics, Speech and Signal Processing (ICASSP)},
  pages={8524--8528},
  year={2020},
  organization={IEEE},
  doi={10.1109/ICASSP40776.2020.9054070}
}

@article{teeter2018generalized,
  title={Generalized Leaky Integrate-and-Fire Models Classify Multiple Neuron Types},
  author={Teeter, Corinne and Iyer, Ramakrishnan and Menon, Vilas and Gouwens, Nathan and Feng, David and Berg, Jim and Szafer, Aaron and others},
  journal={Nature Communications},
  volume={9},
  number={1},
  pages={709},
  year={2018},
  publisher={Nature Publishing Group},
  doi={10.1038/s41467-017-02717-4}
}

@article{poirazi2020illuminating,
  title={Illuminating Dendritic Function with Computational Models},
  author={Poirazi, Panayiota and Papoutsi, Athanasia},
  journal={Nature Reviews Neuroscience},
  volume={21},
  number={6},
  pages={303--321},
  year={2020},
  publisher={Nature Publishing Group},
  doi={10.1038/s41583-020-0301-7}
}

@article{markram2015reconstruction,
  title={Reconstruction and Simulation of Neocortical Microcircuitry},
  author={Markram, Henry and Muller, Eilif and Ramaswamy, Srikanth and Reimann, Michael W. and Abdellah, Marwan and Sanchez, Carlos Aguado and Ailamaki, Anastasia and others},
  journal={Cell},
  volume={163},
  number={2},
  pages={456--492},
  year={2015},
  publisher={Elsevier},
  doi={10.1016/j.cell.2015.09.029}
}

@article{hodgkin1952quantitative,
  title={A Quantitative Description of Membrane Current and its Application to Conduction and Excitation in Nerve},
  author={Hodgkin, Alan L. and Huxley, Andrew F.},
  journal={The Journal of Physiology},
  volume={117},
  number={4},
  pages={500--544},
  year={1952},
  publisher={Wiley-Blackwell},
  doi={10.1113/jphysiol.1952.sp004764}
}

@article{izhikevich2004which,
  title={Which Model to Use for Cortical Spiking Neurons?},
  author={Izhikevich, Eugene M.},
  journal={IEEE Transactions on Neural Networks},
  volume={15},
  number={5},
  pages={1063--1070},
  year={2004},
  publisher={IEEE},
  doi={10.1109/TNN.2004.832719}
}

@article{maass2004computational,
  title={On the Computational Power of Circuits of Spiking Neurons},
  author={Maass, Wolfgang and Markram, Henry},
  journal={Journal of Computer and System Sciences},
  volume={69},
  number={4},
  pages={593--616},
  year={2004},
  publisher={Elsevier},
  doi={10.1016/j.jcss.2004.04.001}
}

@inproceedings{aliyev2024sparsity,
  title={Sparsity-Aware Hardware-Software Co-Design of Spiking Neural Networks: An Overview},
  author={Aliyev, Ilkin and Svoboda, Kama and Adegbija, Tosiron and Fellous, Jean-Marc},
  booktitle={2024 IEEE 17th International Symposium on Embedded Multicore/Many-core Systems-on-Chip (MCSoC)},
  pages={413--420},
  year={2024},
  organization={IEEE}
}

@article{shao2023efficient,
  title={An Efficient Training Accelerator for {Transformers} With Hardware-Algorithm Co-Optimization},
  author={Shao, Haikuo and Lu, Jinming and Wang, Meiqi and Wang, Zhongfeng},
  journal={IEEE Transactions on Very Large Scale Integration (VLSI) Systems},
  volume={31},
  number={11},
  pages={1788--1801},
  year={2023},
  publisher={IEEE},
  doi={10.1109/TVLSI.2023.3312386}
}

@article{svoboda2025spiking,
  title={Spiking Neural Network Architecture Search: A Survey},
  author={Svoboda, Kama and Adegbija, Tosiron},
  journal={arXiv preprint arXiv:2510.14235},
  year={2025},
  doi={10.48550/arXiv.2510.14235}
}

@article{km2025spiking,
  title={Spiking Neural Dynamics and Federated Routing for Ultralow-Latency in {6G} Industry 5.0},
  author={KM, Karthick Raghunath and Babu, TKS Rathish and Albalawi, Eid and TR, Mahesh and Khan, Surbhi Bhatia and Saidani, Oumaima},
  journal={Journal of Sensors},
  volume={2025},
  pages={5491672},
  year={2025},
  publisher={Wiley},
  doi={10.1155/2025/5491672}
}

@inproceedings{huang2024attention,
  title={Attention-Aware Neuromorphic Semantic Communications},
  author={Huang, Haoxiang and Liu, Yanzhen},
  booktitle={2024 IEEE 34th International Workshop on Machine Learning for Signal Processing (MLSP)},
  pages={1--6},
  year={2024},
  organization={IEEE},
  doi={10.1109/MLSP60651.2024.10748197}
}

@article{wang2024snn,
  title={{SNN-SC}: A Spiking Semantic Communication Framework for Collaborative Intelligence},
  author={Wang, Mengyang and Li, Jiahui and Ma, Mengyao and Fan, Xiaopeng},
  journal={IEEE Transactions on Vehicular Technology},
  volume={73},
  number={9},
  pages={12891--12906},
  year={2024},
  publisher={IEEE},
  doi={10.1109/TVT.2024.3406456}
}

@article{carpegna2024spiker,
      title={{Spiker+}: A Framework for the Generation of Efficient Spiking Neural Networks {FPGA} Accelerators for Inference at the Edge},
      author={Carpegna, Alessio and Savino, Alessandro and Di Carlo, Stefano},
      journal={IEEE Transactions on Emerging Topics in Computing},
      year={2024},
      publisher={IEEE},
      doi={10.1109/TETC.2024.3359677},
      note={Early Access}
}

@article{rasmussen2019nengodl,
  title={NengoDL: Combining deep learning and neuromorphic modelling methods},
  author={Rasmussen, Daniel},
  journal={Neuroinformatics},
  volume={17},
  number={4},
  pages={611--628},
  year={2019},
  publisher={Springer},
  doi={10.1007/s12021-019-09424-z}
}

@article{pedersen2024neuromorphic,
      title={Neuromorphic intermediate representation: A unified instruction set for interoperable brain-inspired computing},
      author={Pedersen, Jens E and Abreu, Steven and Jobst, Matthias and Lenz, Gregor and Fra, Vittorio and Bauer, Felix Christian and Muir, Dylan Richard and others},
      journal={Nature Communications},
      volume={15},
      number={1},
      pages={8122},
      year={2024},
      publisher={Nature Publishing Group UK London},
      doi={10.1038/s41467-024-52255-x}
}

@inproceedings{snyder2023neuromorphic,
      title={Neuromorphic {Bayesian} Optimization in {Lava}},
      author={Snyder, Shay and Risbud, Sumedh R and Parsa, Maryam},
      booktitle={Proceedings of the 2023 International Conference on Neuromorphic Systems (ICONS)},
      pages={1--5},
      year={2023},
      publisher={ACM},
      address={New York, NY, USA},
      doi={10.1145/3589737.3605996}
}

@article{zhou2024spikformer,
  title={{Spikformer V2}: Join the High Accuracy Club on {ImageNet} with an {SNN} Ticket},
  author={Zhou, Zhaokun and Che, Kaiwei and Fang, Wei and Tian, Keyu and Zhu, Yuesheng and Yan, Shuicheng and Tian, Yonghong and Yuan, Li},
  journal={arXiv preprint arXiv:2401.02020},
  year={2024},
  doi={10.48550/arXiv.2401.02020}
}

@article{zhu2023spikegpt,
  title={{SpikeGPT}: Generative Pre-trained Language Model with Spiking Neural Networks},
  author={Zhu, Rui-Jie and Zhao, Qihang and Li, Guoqi and Eshraghian, Jason K.},
  journal={arXiv preprint arXiv:2302.13939},
  year={2023},
  doi={10.48550/arXiv.2302.13939}
}

@inproceedings{NEURIPS2024_7c9341ad,
 author = {Hwang, Sangwoo and Lee, Seunghyun and Park, Dahoon and Lee, Donghun and Kung, Jaeha},
 booktitle = {Advances in Neural Information Processing Systems},
 doi = {10.52202/079017-2152},
 editor = {A. Globerson and L. Mackey and D. Belgrave and A. Fan and U. Paquet and J. Tomczak and C. Zhang},
 pages = {67422--67445},
 publisher = {Curran Associates, Inc.},
 title = {SpikedAttention: Training-Free and Fully Spike-Driven Transformer-to-SNN Conversion with Winner-Oriented Spike Shift for Softmax Operation},
 volume = {37},
 year = {2024}
}

@article{bu2023optimal,
  title={Optimal {ANN-SNN} Conversion for High-accuracy and Ultra-low-latency Spiking Neural Networks},
  author={Bu, Tong and Fang, Wei and Ding, Jianhao and Dai, Penglin and Yu, Zhaofei and Huang, Tiejun},
  journal={arXiv preprint arXiv:2303.04347},
  year={2023},
  doi={10.48550/arXiv.2303.04347}
}

@inproceedings{NEURIPS2022_82846e19,
 author = {Xiao, Mingqing and Meng, Qingyan and Zhang, Zongpeng and He, Di and Lin, Zhouchen},
 booktitle = {Advances in Neural Information Processing Systems},
 editor = {S. Koyejo and S. Mohamed and A. Agarwal and D. Belgrave and K. Cho and A. Oh},
 pages = {20717--20730},
 publisher = {Curran Associates, Inc.},
 title = {Online Training Through Time for Spiking Neural Networks},
 volume = {35},
 year = {2022}
}

@article{Ref_Existing_General_Reviews,
author = {GHOSH-DASTIDAR, SAMANWOY and ADELI, HOJJAT},
title = {SPIKING NEURAL NETWORKS},
journal = {International Journal of Neural Systems},
volume = {19},
number = {04},
pages = {295-308},
year = {2009},
doi = {10.1142/S0129065709002002}
}

@article{schuman2022opportunities,
  title={Opportunities for neuromorphic computing algorithms and applications},
  author={Schuman, Catherine D and Kulkarni, Shuman R and Parsa, Maryam and Mitchell, J Parker and Date, Prasanna and Kay, Bill},
  journal={Nature Computational Science},
  volume={2},
  number={1},
  pages={10--19},
  year={2022},
  publisher={Nature Publishing Group}
}

@ARTICLE{Ref_Neuromorphic_Hardware1,
  author={Li, Guoqi and Deng, Lei and Tang, Huajin and Pan, Gang and Tian, Yonghong and Roy, Kaushik and Maass, Wolfgang},
  journal={Proceedings of the IEEE}, 
  title={Brain-Inspired Computing: A Systematic Survey and Future Trends}, 
  year={2024},
  volume={112},
  number={6},
  pages={544-584},
  doi={10.1109/JPROC.2024.3429360}}

@INPROCEEDINGS{Ref_Neuromorphic_Hardware2,
  author={Meyer, Svea Marie and Weidel, Philipp and Plank, Philipp and Campos-Macias, Leobardo and Shreshta, Sumit Bam and Stratmann, Philipp and Timcheck, Jonathan and Richter, Mathis},
  booktitle={2025 Neuro Inspired Computational Elements (NICE)}, 
  title={A Diagonal Structured State Space Model on Loihi 2 for Efficient Streaming Sequence Processing}, 
  year={2025},
  volume={},
  number={},
  pages={1-9},
  doi={10.1109/NICE65350.2025.11065663}}

@INPROCEEDINGS{Ref_Neuromorphic_Hardware3,
  author={Varadarajulu, Swetha and Mendonça, Marcele O. K. and Eappen, Geoffrey and Querol, Jorge and Chatzinotas, Symeon},
  booktitle={41st International Communications Satellite Systems Conference (ICSSC 2024)}, 
  title={Enhanced demodulator for 5G NTN using spatio-temporal attention convolutional autoencoder and Akida Brainchip SNN}, 
  year={2024},
  volume={2024},
  number={},
  pages={99-104},
  doi={10.1049/icp.2024.4619}}

@article{Ref_COTS_SNN_Inefficiency3,
  title={A High Energy-Efficiency Multi-core Neuromorphic Architecture for {Deep SNN} Training},
  author={Li, Mingjing and Ding, Jing and Yue, Zheyu and Li, Yuhang and Luo, Zuoheng and Fan, Zhaozhi and Zhao, Wenjia and Min, Haolin and Liu, Fangxin and Yang, Xiaokang and Liang, Xiaoyao and Jiang, Li},
  journal={arXiv preprint arXiv:2412.05302},
  year={2024},
  doi={10.48550/arXiv.2412.05302}
}

@article{Ref_COTS_SNN_Inefficiency2,
  title={Reconsidering the Energy Efficiency of {Spiking Neural Networks}},
  author={Yan, Zhanglu and Bai, Zhenyu and Tang, Kaiwen and Wong, Weng-Fai},
  journal={arXiv preprint arXiv:2409.08290},
  year={2024},
  doi={10.48550/arXiv.2409.08290}
}

@ARTICLE{Ref_SNN_Sparsity_Efficiency4,
  author={Chu, Haoming and Yan, Yulong and Gan, Leijing and Jia, Hao and Qian, Liyu and Huan, Yuxiang and Zheng, Lirong and Zou, Zhuo},
  journal={IEEE Transactions on Biomedical Circuits and Systems}, 
  title={A Neuromorphic Processing System With Spike-Driven SNN Processor for Wearable ECG Classification}, 
  year={2022},
  volume={16},
  number={4},
  pages={511-523},
  doi={10.1109/TBCAS.2022.3189364}}

@ARTICLE{Ref_SNN_Sparsity_Efficiency3,
  author={Kim, Sangyeob and Kim, Soyeon and Hong, Seongyon and Kim, Sangjin and Han, Donghyeon and Choi, Jiwon and Yoo, Hoi-Jun},
  journal={IEEE Journal of Solid-State Circuits}, 
  title={C-DNN: An Energy-Efficient Complementary Deep-Neural-Network Processor With Heterogeneous CNN/SNN Core Architecture}, 
  year={2024},
  volume={59},
  number={1},
  pages={157-172},
  doi={10.1109/JSSC.2023.3330483}}

@ARTICLE{Ref_SNN_Sparsity_Efficiency2,
  author={Liu, Fangxin and Wang, Zongwu and Zhao, Wenbo and Yang, Ning and Chen, Yongbiao and Huang, Shiyuan and Li, Haomin and Yang, Tao and Pei, Songwen and Liang, Xiaoyao and Jiang, Li},
  journal={IEEE Transactions on Parallel and Distributed Systems}, 
  title={Exploiting Temporal-Unrolled Parallelism for Energy-Efficient SNN Acceleration}, 
  year={2024},
  volume={35},
  number={10},
  pages={1749-1764},
  doi={10.1109/TPDS.2024.3415712}}

@ARTICLE{Ref_SNN_Sparsity_Efficiency1,
  author={Fang, Chaoming and Shen, Ziyang and Wang, Zongsheng and Wang, Chuanqing and Zhao, Shiqi and Tian, Fengshi and Yang, Jie and Sawan, Mohamad},
  journal={IEEE Journal of Solid-State Circuits}, 
  title={An Energy-Efficient Unstructured Sparsity-Aware Deep SNN Accelerator With 3-D Computation Array}, 
  year={2025},
  volume={60},
  number={3},
  pages={977-989},
  doi={10.1109/JSSC.2024.3507095}}

@INPROCEEDINGS{Ref_VonNeumann_EdgeAI1,
  author={Dong, Fangyuan and Si, Xin and Chang, Meng-Fan},
  booktitle={2022 IEEE 16th International Conference on Solid-State \& Integrated Circuit Technology (ICSICT)}, 
  title={Design Methodology and Trends of SRAM-Based Compute-in-Memory Circuits}, 
  year={2022},
  volume={},
  number={},
  pages={1-4}}

@ARTICLE{Ref_VonNeumann_EdgeAI2,
  author={Ramírez, Cristian and Castelló, Adrián and Martínez, Héctor and Quintana-Ortí, Enrique S.},
  journal={IEEE Internet of Things Journal}, 
  title={Communication-Avoiding Fusion of GEMM-Based Convolutions for Deep Learning in the RISC-V GAP8 MCU}, 
  year={2024},
  volume={11},
  number={21},
  pages={35640-35653},
  doi={10.1109/JIOT.2024.3436937}}

@ARTICLE{Ref_VonNeumann_EdgeAI3,
  author={Sun, Yuting and Chen, Tong and Nguyen, Quoc Viet Hung and Yin, Hongzhi},
  journal={IEEE Transactions on Industrial Informatics}, 
  title={TinyAD: Memory-Efficient Anomaly Detection for Time-Series Data in Industrial IoT}, 
  year={2024},
  volume={20},
  number={1},
  pages={824-834},
  doi={10.1109/TII.2023.3254668}}

@article{Imam2020Olfaction,
  title={Rapid online learning and robust recall in a neuromorphic olfactory circuit},
  author={Imam, Nabil and Cleland, Thomas A},
  journal={Nature Machine Intelligence},
  volume={2},
  number={3},
  pages={181--191},
  year={2020},
  publisher={Nature Publishing Group}
}

@article{Yin2021Auditory,
  title={Accurate and efficient time-domain classification with adaptive spiking recurrent neural networks},
  author={Yin, Bojian and Corradi, Federico and Boht{\'e}, Sander M},
  journal={Nature Machine Intelligence},
  volume={3},
  number={10},
  pages={905--913},
  year={2021},
  publisher={Nature Publishing Group}
}

@article{Wang2024Gas,
  title={Anti-Drift Gas Detection Algorithm Based on Neural Network},
  author={Guo, Jiayi and Li, Xu and Li, Xiulei and Liang, Zheng and Cao, Juexian and Wei, Xiaolin},
  journal={IEEE Transactions on Instrumentation and Measurement},
  year={2024},
  publisher={IEEE}
}

@article{Liang2021Exploring,
  title={Exploring adversarial attack in spiking neural networks with spike-compatible gradient},
  author={Liang, Ling and Hu, Xing and Deng, Lei and Wu, Yujie and Li, Guoqi and Ding, Yufei and Li, Peng and Xie, Yuan},
  journal={IEEE Transactions on Neural Networks and Learning Systems (TNNLS)},
  volume={34},
  number={5},
  pages={2569--2583},
  year={2023},
  publisher={IEEE}
}

@article{Huang2021SideChannel,
  title={Neural network structure through remote FPGA side-channel analysis},
  author={Huang, Qingrong and Yang, Zeyu and Ni, Kai and Imani, Mohsen and Zhuo, Cheng and Yu, Shimeng and Koushanfar, Farinaz},
  journal={IEEE Transactions on Information Forensics and Security (TIFS)},
  volume={16},
  pages={4377--4388},
  year={2021},
  publisher={IEEE}
}

@inproceedings{Na2022AutoSNN,
  title={Autosnn: Towards energy-efficient spiking neural networks},
  author={Na, Byunggook and Mok, Jisoo and Park, Seongsik and Lee, Dongjin and Choe, Hyeokjun and Yoon, Sungroh},
  booktitle={International conference on machine learning},
  pages={16253--16269},
  year={2022},
  organization={PMLR}
}

@inproceedings{kim2022neural,
  title={Neural architecture search for spiking neural networks},
  author={Kim, Youngeun and Li, Yuhang and Park, Hyoungseob and Venkatesha, Yeshwanth and Panda, Priyadarshini},
  booktitle={European conference on computer vision},
  pages={36--56},
  year={2022},
  organization={Springer}
}

@inproceedings{Che2022SpikeDHS,
  title={Differentiable Hierarchical and Surrogate Gradient Search for Spiking Neural Networks},
  author={Che, Kaiwei and Leng, Luziwei and Zhang, Kaixuan and Zhang, Jianguo and Cheng, Jian},
  booktitle={Advances in Neural Information Processing Systems (NeurIPS)},
  volume={35},
  pages={24975--24990},
  year={2022}
}

@article{Yik2024Neuroevolution,
  title={Neuroevolutionary Learning for Hardware-Aware Spiking Neural Networks},
  author={Yik, Jason and Ahmed, Zishen and Anderson, Andrew G.},
  journal={IEEE Transactions on Pattern Analysis and Machine Intelligence (TPAMI)},
  volume={46},
  number={5},
  pages={2845--2858},
  year={2024},
  publisher={IEEE}
}

@inproceedings{Li2021S2N2,
  title={S2N2: A FPGA accelerator for streaming spiking neural networks},
  author={Li, Alireza and Liu, Y. and Kastner, Ryan},
  booktitle={Proceedings of the 2021 ACM/SIGDA International Symposium on Field-Programmable Gate Arrays (FPGA)},
  pages={1--11},
  year={2021}
}

@INPROCEEDINGS{Panchapakesan2021,
  author={Panchapakesan, Sathish and Fang, Zhenman and Li, Jian},
  booktitle={2021 31st International Conference on Field-Programmable Logic and Applications (FPL)}, 
  title={SyncNN: Evaluating and Accelerating Spiking Neural Networks on FPGAs}, 
  year={2021},
  volume={},
  number={},
  pages={286-293},
  doi={10.1109/FPL53798.2021.00058}}

@article{Kang2023TCAS,
  title={A 24.3 $\mu$J/Image SNN Accelerator for DVS-Gesture With WS-LOS Dataflow and Sparse Methods},
  author={Kang, Lei and Yang, Xu},
  journal={IEEE Transactions on Circuits and Systems II: Express Briefs},
  volume={70},
  number={10},
  pages={3802--3806},
  year={2023},
  publisher={IEEE}
}

@article{FireFlyV2_2024,
  title={FireFly v2: Advancing Hardware Support for High-Performance Spiking Neural Network With a Spatiotemporal FPGA Accelerator},
  author={Li, J. and others},
  journal={IEEE Transactions on Computer-Aided Design of Integrated Circuits and Systems (TCAD)},
  volume={43},
  number={9},
  pages={1-14},
  year={2024},
  publisher={IEEE}
}

@article{SpikeX2025,
  title={SpikeX: Exploring Accelerator Architecture and Network-Hardware Co-Optimization for Sparse Spiking Neural Networks},
  author={Xu, Boxun and others},
  journal={IEEE Transactions on Computer-Aided Design of Integrated Circuits and Systems (TCAD)},
  note={(Early Access / arXiv:2505.12292)},
  year={2025},
  publisher={IEEE}
}

@article{Maass1997,
  title={Networks of spiking neurons: The third generation of neural network models},
  author={Maass, Wolfgang},
  journal={Neural Networks},
  volume={10},
  number={9},
  pages={1659--1671},
  year={1997},
  publisher={Elsevier}
}

@book{Gerstner2002,
  title={Spiking neuron models: Single neurons, populations, plasticity},
  author={Gerstner, Wulfram and Kistler, Werner M},
  year={2002},
  publisher={Cambridge University Press}
}

@article{Izhikevich2003,
  title={Simple model of spiking neurons},
  author={Izhikevich, Eugene M},
  journal={IEEE Transactions on Neural Networks},
  volume={14},
  number={6},
  pages={1569--1572},
  year={2003},
  publisher={IEEE}
}

@article{Eshraghian2023Survey,
  title={Training spiking neural networks using lessons from deep learning},
  author={Eshraghian, Jason K and Ward, Max and Neftci, Emre and others},
  journal={Proceedings of the IEEE},
  volume={111},
  number={9},
  pages={1016--1054},
  year={2023},
  publisher={IEEE}
}

@article{Ma2024Darwin3,
  title={Darwin3: A large-scale neuromorphic chip with a novel ISA and on-chip learning},
  author={Ma, De and Jin, Xiaofei and Sun, Shichun and Li, Yitao and Wu, Xundong and Hu, Youneng and Yang, Fangchao and Tang, Huajin and Pan, Gang},
  journal={National Science Review},
  volume={11},
  number={5},
  pages={nwae102},
  year={2024},
  publisher={Oxford University Press}
}

@article{Ma2017Darwin,
  title={Darwin: A neuromorphic hardware co-processor based on spiking neural networks},
  author={Ma, De and Shen, Juncheng and Gu, Zonghua and Zhang, Ming and Zhu, Xiaolei and Xu, Qiang and Shen, Qi and Pan, Gang},
  journal={Journal of Systems Architecture},
  volume={77},
  pages={43--51},
  year={2017},
  publisher={Elsevier}
}

@article{Huang2022Vidar,
  title={Ultra-high temporal resolution visual reconstruction from a fovea-like spike camera},
  author={Zhu, Lin and Dong, Siwei and Huang, Tiejun and Tian, Yonghong},
  journal={IEEE Transactions on Pattern Analysis and Machine Intelligence (TPAMI)},
  volume={44},
  number={11},
  pages={7785--7799},
  year={2022},
  publisher={IEEE}
}

@article{Dong2024SpikeCam,
  title={Spike cameras: A review},
  author={Dong, Siwei and Zhu, Lin and Duan, Da and Huang, Tiejun and Tian, Yonghong},
  journal={arXiv preprint arXiv:2401.13490},
  year={2024}
}

@article{Pei2019Tianjic,
  title={Towards artificial general intelligence with hybrid Tianjic chip architecture},
  author={Pei, Jing and Deng, Lei and Song, Sen and Zhao, Mingguo and Zhang, Youhui and Wu, Shuang and Wang, Guanrui and Zou, Zhe and Wu, Zhenzhi and He, Wei and others},
  journal={Nature},
  volume={572},
  number={7767},
  pages={206--211},
  year={2019},
  publisher={Nature Publishing Group}
}

@article{Wu2018STBP,
  title={Spatio-temporal backpropagation for training high-performance spiking neural networks},
  author={Wu, Yujie and Deng, Lei and Li, Guoqi and Zhu, Jun and Shi, Luping},
  journal={Frontiers in Neuroscience},
  volume={12},
  pages={331},
  year={2018},
  publisher={Frontiers}
}

@article{Wu2019STBP_PAMI,
  title={Direct training for spiking neural networks: Faster, larger, and better},
  author={Wu, Yujie and Deng, Lei and Li, Guoqi and Zhu, Jun and Shi, Luping},
  journal={Proceedings of the AAAI Conference on Artificial Intelligence},
  volume={33},
  number={01},
  pages={1311--1318},
  year={2019}
}

@inproceedings{Zheng2021tdBN,
  title={Going deeper with directly-trained larger spiking neural networks},
  author={Zheng, Hanle and Wu, Yujie and Deng, Lei and Hu, Yifan and Li, Guoqi},
  booktitle={Proceedings of the AAAI Conference on Artificial Intelligence},
  volume={35},
  number={12},
  pages={11062--11070},
  year={2021}
}

@article{karamimanesh2025fpga,
  title={Spiking Neural Networks on {FPGA}: A Survey of Methodologies and Recent Advancements},
  author={Karamimanesh, Mehrzad and Abiri, Ebrahim and Shahsavari, Mahyar and Hassanli, Kourosh and van Schaik, Andr{\'e} and Eshraghian, Jason},
  journal={Neural Networks},
  pages={107256},
  year={2025},
  publisher={Elsevier}
}

@inproceedings{Liang2025,
  title={Towards Accurate Binary Spiking Neural Networks: Learning with Adaptive Gradient Modulation Mechanism},
  author={Liang, Yu and Wei, Wenjie and Belatreche, Ammar and Cao, Honglin and Zhou, Zijian and Wang, Shuai and Zhang, Malu and Yang, Yang},
  booktitle={Proceedings of the AAAI Conference on Artificial Intelligence},
  volume={39},
  number={2},
  pages={1402--1410},
  year={2025}
}

@article{Yamazaki2022,
  title={Spiking Neural Networks and Their Applications: A Review},
  author={Yamazaki, Kashu and Vo-Ho, Viet-Khoa and Bulsara, Darshan and Le, Ngan},
  journal={Brain Sciences},
  volume={12},
  number={7},
  pages={863},
  year={2022},
  publisher={MDPI}
}

@ARTICLE{Iaboni2024,
  author={Iaboni, Craig and Abichandani, Pramod},
  journal={IEEE Access}, 
  title={Event-Based Spiking Neural Networks for Object Detection: A Review of Datasets, Architectures, Learning Rules, and Implementation}, 
  year={2024},
  volume={12},
  number={},
  pages={180532-180596},
  doi={10.1109/ACCESS.2024.3479968}}

@article{ZhangGuanlei2025,
  author={Zhang, Guanlei and Feng, Lei and Zhou, Fanqin and Yang, Zhixiang and Zhang, Qiyang and Saleh, Alaa and Donta, Praveen Kumar and Dehury, Chinmaya Kumar},
  journal={IEEE Consumer Electronics Magazine}, 
  title={Spiking Neural Networks in Intelligent Edge Computing}, 
  year={2025},
  volume={14},
  number={4},
  pages={66-75},
  doi={10.1109/MCE.2024.3506502}
}

@ARTICLE{HuoBingqiang2025,
  author={Huo, Bingqiang and Li, Fang and Peng, Siyu and Chen, Hongwei and Xin, Sudan and Wang, Hongjun},
  journal={IEEE Access}, 
  title={Research on SNN Learning Algorithms and Networks Based on Biological Plausibility}, 
  year={2025},
  volume={13},
  number={},
  pages={95243-95256}}

@article{ShenShuaijie2024,
  title={Evolutionary Spiking Neural Networks: A Survey},
  author={Shen, Shuaijie and Zhang, Rui and Wang, Chao and Huang, Renzhuo and Tuerhong, Aiersi and Guo, Qinghai and Lu, Zhichao and Zhang, Jianguo and Leng, Luziwei},
  journal={Journal of Membrane Computing},
  pages={1--12},
  year={2024},
  publisher={Springer}
}

@article{Malcolm2023,
  title={A Comprehensive Review of Spiking Neural Networks: Interpretation, Optimization, Efficiency, and Best Practices},
  author={Malcolm, Kai and Casco-Rodriguez, Josue},
  journal={arXiv preprint arXiv:2303.10780},
  year={2023}
}

@article{XiaoChao2022,
  title={Optimal Mapping of Spiking Neural Network to Neuromorphic Hardware for {Edge-AI}},
  author={Xiao, Chao and Chen, Jihua and Wang, Lei},
  journal={Sensors},
  volume={22},
  number={19},
  pages={7248},
  year={2022},
  publisher={MDPI}
}

@article{WeiWenjie2024,
  title={Event-Driven Learning for Spiking Neural Networks},
  author={Wei, Wenjie and Zhang, Malu and Zhang, Jilin and Belatreche, Ammar and Wu, Jibin and Xu, Zijing and Qiu, Xuerui and Chen, Hong and Yang, Yang and Li, Haizhou},
  journal={arXiv preprint arXiv:2403.00270},
  year={2024}
}

@article{Rathi2023,
  author = {Rathi, Nitin and Chakraborty, Indranil and Kosta, Adarsh and Sengupta, Abhronil and Ankit, Aayush and Panda, Priyadarshini and Roy, Kaushik},
  title = {Exploring Neuromorphic Computing Based on Spiking Neural Networks: Algorithms to Hardware},
  year = {2023},
  issue_date = {December 2023},
  publisher = {Association for Computing Machinery},
  address = {New York, NY, USA},
  volume = {55},
  number = {12},
  issn = {0360-0300},
  doi = {10.1145/3571155},
  journal = {ACM Computing Surveys},
  month = mar,
  articleno = {243},
  numpages = {49}
}

@article{JiangJiaqiang2025,
  title={Adaptive Gradient Learning for Spiking Neural Networks by Exploiting Membrane Potential Dynamics},
  author={Jiang, Jiaqiang and Wang, Lei and Jiang, Runhao and Fan, Jing and Yan, Rui},
  journal={arXiv preprint arXiv:2505.11863},
  year={2025}
}

@article{LiSai2025,
  title={{QUEST}: A Quantized Energy-Aware {SNN} Training Framework for Multi-State Neuromorphic Devices},
  author={Li, Sai and Chen, Linliang and Zhang, Yihao and Zhang, Zhongkui and Du, Ao and Pan, Biao and Wang, Zhaohao and Wen, Lianggong and Zhao, Weisheng},
  journal={arXiv preprint arXiv:2504.00679},
  year={2025}
}

@article{LiZiwen2024,
author = {Li, Ziwen and Ma, Yu and Zhou, Jindong and Zhou, Pingqiang},
title = {Spiking-NeRF: Spiking Neural Network for Energy-Efficient Neural Rendering},
year = {2024},
issue_date = {July 2024},
publisher = {Association for Computing Machinery},
address = {New York, NY, USA},
volume = {20},
number = {3},
issn = {1550-4832},
doi = {10.1145/3675808},
journal = {J. Emerg. Technol. Comput. Syst.},
month = aug,
articleno = {10},
numpages = {23}
}

@InProceedings{Ahmed2025,
    author    = {Ahmed, Soikat Hasan and Finkbeiner, Jan and Neftci, Emre},
    title     = {Efficient Event-Based Object Detection: A Hybrid Neural Network with Spatial and Temporal Attention},
    booktitle = {Proceedings of the IEEE/CVF Conference on Computer Vision and Pattern Recognition (CVPR)},
    month     = {June},
    year      = {2025},
    pages     = {13970-13979}
}

@article{ZhouChenlin2024,
  title={Direct Training High-Performance Deep Spiking Neural Networks: A Review of Theories and Methods},
  author={Zhou, Chenlin and Zhang, Han and Yu, Liutao and Ye, Yumin and Zhou, Zhaokun and Huang, Liwei and Ma, Zhengyu and Fan, Xiaopeng and Zhou, Huihui and Tian, Yonghong},
  journal={Frontiers in Neuroscience},
  volume={18},
  pages={1383844},
  year={2024},
  publisher={Frontiers Media SA}
}

@article{HaoZecheng2024,
  title={Faster and Stronger: When {ANN-SNN} Conversion Meets Parallel Spiking Calculation},
  author={Hao, Zecheng and Yu, Zhaofei and Huang, Tiejun},
  journal={arXiv preprint arXiv:2412.13610},
  year={2024}
}

@inproceedings{YuDi2025,
  title     = {ECC-SNN: Cost-Effective Edge-Cloud Collaboration for Spiking Neural Networks},
  author    = {Yu, Di and Lv, Changze and Du, Xin and Jiang, Linshan and Tong, Wentao and Liao, Zhenyu and Zheng, Xiaoqing and Deng, Shuiguang},
  booktitle = {Proceedings of the Thirty-Fourth International Joint Conference on Artificial Intelligence (IJCAI-25)},
  publisher = {International Joint Conferences on Artificial Intelligence Organization},
  editor    = {Kwok, James},
  pages     = {6904--6912},
  year      = {2025},
  month     = aug,
  doi       = {10.24963/ijcai.2025/768}
}

@inproceedings{Aydin2024,
  title={A Hybrid {ANN-SNN} Architecture for Low-Power and Low-Latency Visual Perception},
  author={Aydin, Asude and Gehrig, Mathias and Gehrig, Daniel and Scaramuzza, Davide},
  booktitle={Proceedings of the IEEE/CVF Conference on Computer Vision and Pattern Recognition (CVPR)},
  pages={5701--5711},
  year={2024}
}

@inproceedings{Seekings2024,
  title={Towards Efficient Deployment of Hybrid {SNNs} on Neuromorphic and Edge {AI} Hardware},
  author={Seekings, James and Chandarana, Peyton and Ardakani, Mahsa and Mohammadi, MohammadReza and Zand, Ramtin},
  booktitle={2024 International Conference on Neuromorphic Systems (ICONS)},
  pages={71--77},
  year={2024},
  organization={IEEE}
}

@article{Rivelli2025,
  title={Adaptively Pruned Spiking Neural Networks for Energy-Efficient Intracortical Neural Decoding},
  author={Rivelli, Francesca and Popov, Martin and Kouzinopoulos, Charalampos S and Tang, Guangzhi},
  journal={arXiv preprint arXiv:2504.11568},
  year={2025}
}

@inproceedings{Schmolli2025,
  title     = {Adversarially Robust Spiking Neural Networks with Sparse Connectivity},
  author    = {Schmolli, Mathias and Baronig, Maximilian and Legenstein, Robert and Ozdenizci, Ozan},
  booktitle = {Proceedings of the Conference on Parsimony and Learning (CPAL)},
  series    = {Proceedings of Machine Learning Research},
  volume    = {280},
  pages     = {865--883},
  year      = {2025},
  editor    = {Chen, Beidi and Liu, Shijia and Pilanci, Mert and Su, Weijie and Sulam, Jeremias and Wang, Yuxiang and Zhu, Zhihui},
  publisher = {PMLR},
  month     = mar
}

@article{ShiLianfeng2025,
  title     = {Optimal Spiking Brain Compression: Improving One-Shot Post-Training Pruning and Quantization for Spiking Neural Networks},
  author    = {Shi, Lianfeng and Li, Ao and Ward-Cherrier, Benjamin},
  journal   = {arXiv preprint arXiv:2506.03996},
  year      = {2025},
  month     = jun,
  doi       = {10.48550/arXiv.2506.03996}
}

@article{PutraRachmad2024,
  title={{SNN4Agents}: A Framework for Developing Energy-Efficient Embodied Spiking Neural Networks for Autonomous Agents},
  author={Putra, Rachmad Vidya Wicaksana and Marchisio, Alberto and Shafique, Muhammad},
  journal={Frontiers in Robotics and AI},
  volume={11},
  pages={1401677},
  year={2024},
  publisher={Frontiers Media SA}
}

@article{Vogginger2024,
  title={Neuromorphic Hardware for Sustainable {AI} Data Centers},
  author={Vogginger, Bernhard and Rostami, Amirhossein and Jain, Vaibhav and Arfa, Sirine and Hantsch, Andreas and Kappel, David and Sch{\"a}fer, Michael and Faltings, Ulrike and Gonzalez, Hector A and Liu, Chen and others},
  journal={arXiv preprint arXiv:2402.02521},
  year={2024}
}

@article{WangHuan2023,
  title={Brain-Inspired Spiking Neural Networks for Industrial Fault Diagnosis: A Survey, Challenges, and Opportunities},
  author={Wang, Huan and Li, Yan-Fu and Gryllias, Konstantinos},
  journal={arXiv preprint arXiv:2401.02429},
  year={2023}
}

@article{XueJianwei2023,
  title={{EdgeMap}: An Optimized Mapping Toolchain for Spiking Neural Network in Edge Computing},
  author={Xue, Jianwei and Xie, Lisheng and Chen, Faquan and Wu, Liangshun and Tian, Qingyang and Zhou, Yifan and Ying, Rendong and Liu, Peilin},
  journal={Sensors},
  volume={23},
  number={14},
  pages={6548},
  year={2023},
  publisher={MDPI}
}

@INPROCEEDINGS{GomeWalter2023,
  author={Gomez, Walter Gallego and Pignata, Andrea and Pignari, Riccardo and Fra, Vittorio and Macii, Enrico and Urgese, Gianvito},
  booktitle={2023 IEEE 16th International Symposium on Embedded Multicore/Many-core Systems-on-Chip (MCSoC)}, 
  title={First Steps Towards Micro-Benchmarking the {Lava-Loihi} Neuromorphic Ecosystem}, 
  year={2023},
  pages={462-469},
  doi={10.1109/MCSoC60832.2023.00075}
}

@ARTICLE{LiuYanzhen2024,
  author={Liu, Yanzhen and Qin, Zhijin and Li, Geoffrey Ye},
  journal={IEEE Transactions on Wireless Communications}, 
  title={Energy-Efficient Distributed Spiking Neural Network for Wireless Edge Intelligence}, 
  year={2024},
  volume={23},
  number={9},
  pages={10683-10697},
  doi={10.1109/TWC.2024.3374549}
}

@article{YuDongfang2024,
  title={{FedLEC}: Effective Federated Learning Algorithm with Spiking Neural Networks Under Label Skews},
  author={Yu, Dongfang and Du, Xu and Jiang, Long and Li, Yuan},
  journal={arXiv preprint arXiv:2404.14488},
  year={2024}
}

@article{PutraRachmad2025,
  title={Enabling Efficient Processing of Spiking Neural Networks with On-Chip Learning on Commodity Neuromorphic Processors for Edge {AI} Systems},
  author={Putra, Rachmad Vidya Wicaksana and Wickramasinghe, Pasindu and Shafique, Muhammad},
  journal={arXiv preprint arXiv:2504.00957},
  year={2025}
}

@ARTICLE{Hussaini2025,
  author={Hussaini, Somayeh and Milford, Michael and Fischer, Tobias},
  journal={IEEE Transactions on Robotics}, 
  title={Applications of Spiking Neural Networks in Visual Place Recognition}, 
  year={2025},
  volume={41},
  pages={518-537},
  doi={10.1109/TRO.2024.3508053}
}

@article{SabbellaHemanth2025,
  title={The Promise of Spiking Neural Networks for Ubiquitous Computing: A Survey and New Perspectives},
  author={Sabbella, Hemanth and Mukherjee, Archit and Kandappu, Thivya and Dey, Sounak and Pal, Arpan and Misra, Archan and Ma, Dong},
  journal={arXiv preprint arXiv:2506.01737},
  year={2025}
}

@article{ZhengNaichuan2025,
  title={{SNN}-Driven Multimodal Human Action Recognition via Event Camera and Skeleton Data Fusion},
  author={Zheng, Naichuan and Xia, Hailun},
  journal={arXiv preprint arXiv:2502.13385},
  year={2025}
}

@article{BaekSuwhan2024,
  title={{SNN} and Sound: A Comprehensive Review of Spiking Neural Networks in Sound},
  author={Baek, Suwhan and Lee, Jaewon},
  journal={Biomedical Engineering Letters},
  volume={14},
  number={5},
  pages={981--991},
  year={2024},
  publisher={Springer}
}

@article{ChoiHansol2024,
  title={Review on Spiking Neural Network-Based {ECG} Classification Methods for Low-Power Environments},
  author={Choi, Hansol and Park, Jangsoo and Lee, Jongseok and Sim, Donggyu},
  journal={Biomedical Engineering Letters},
  volume={14},
  number={5},
  pages={917--941},
  year={2024},
  publisher={Springer}
}

@article{HizemMoez2025,
  title={Reliable {ECG} Anomaly Detection on Edge Devices for {Internet of Medical Things} Applications},
  author={Hizem, Moez and Bousbia, Leila and Ben Dhiab, Yassmine and Aoueileyine, Mohamed Ould-Elhassen and Bouallegue, Ridha},
  journal={Sensors},
  volume={25},
  number={8},
  pages={2496},
  year={2025},
  publisher={MDPI}
}

@article{Alzarooni2025,
  title={Anomaly Detection for Industrial Applications, Its Challenges, Solutions, and Future Directions: A Review},
  author={Alzarooni, Abdelrahman and Iqbal, Ehtesham and Khan, Samee Ullah and Javed, Sajid and Moyo, Brain and Abdulrahman, Yusra},
  journal={arXiv preprint arXiv:2501.11310},
  year={2025}
}

@article{ZhouDeming2023,
  title={Fast and Accurate {SNN} Model Strengthening for Industrial Applications},
  author={Zhou, Deming and Chen, Weitong and Chen, Kongyang and Mi, Bing},
  journal={Electronics},
  volume={12},
  number={18},
  pages={3845},
  year={2023},
  publisher={MDPI}
}

@article{DolatAbadi2025,
  title={Revolutionizing Traffic Management with {AI-Powered} Machine Vision: A Step Toward Smart Cities},
  author={DolatAbadi, Seyed Hossein Hosseini and Hashemi, Sayyed Mohammad Hossein and Hosseini, Mohammad and AliHosseini, Moein-Aldin},
  journal={arXiv preprint arXiv:2503.02967},
  year={2025}
}

@article{PalOsim2023,
  title={A Comprehensive Review of {AI}-Enabled Unmanned Aerial Vehicle: Trends, Vision, and Challenges},
  author={Pal, Osim Kumar and Shovon, Md Sakib Hossain and Mridha, Muhammad Firoz and Shin, Jungpil},
  journal={arXiv preprint arXiv:2310.16360},
  year={2023}
}

@article{Wolniak2024,
  title={Artificial Intelligence in Smart Cities—Applications, Barriers, and Future Directions: A Review},
  author={Wolniak, Rados{\l}aw and Stecu{\l}a, Kinga},
  journal={Smart Cities},
  volume={7},
  number={3},
  pages={1346--1389},
  year={2024},
  publisher={MDPI}
}

@article{PuglieseViloria2024,
  title={Hazard Susceptibility Mapping with Machine and Deep Learning: A Literature Review},
  author={Pugliese Viloria, Angelly de Jesus and Folini, Andrea and Carrion, Daniela and Brovelli, Maria Antonia},
  journal={Remote Sensing},
  volume={16},
  number={18},
  pages={3374},
  year={2024},
  publisher={MDPI}
}

@article{WangXubin2025,
  title={Empowering Edge Intelligence: A Comprehensive Survey on On-Device {AI} Models},
  author={Wang, Xubin and Tang, Zhiqing and Guo, Jianxiong and Meng, Tianhui and Wang, Chenhao and Wang, Tian and Jia, Weijia},
  journal={ACM Computing Surveys},
  volume={57},
  number={9},
  pages={1--39},
  year={2025},
  publisher={ACM New York, NY}
}

@ARTICLE{WangChengXiang2023,
  author={Wang, Cheng-Xiang and You, Xiaohu and Gao, Xiqi and Zhu, Xiuming and Li, Zixin and Zhang, Chuan and Wang, Haiming and Huang, Yongming and Chen, Yunfei and Haas, Harald and Thompson, John S. and Larsson, Erik G. and Renzo, Marco Di and Tong, Wen and Zhu, Peiying and Shen, Xuemin and Poor, H. Vincent and Hanzo, Lajos},
  journal={IEEE Communications Surveys \& Tutorials}, 
  title={On the Road to {6G}: Visions, Requirements, Key Technologies, and Testbeds}, 
  year={2023},
  volume={25},
  number={2},
  pages={905-974},
  doi={10.1109/COMST.2023.3249835}
}

@article{MaassWolfgang2015,
  title={To Spike or Not to Spike: That is the Question},
  author={Maass, Wolfgang},
  journal={Proceedings of the IEEE},
  volume={103},
  number={12},
  pages={2219--2224},
  year={2015},
  publisher={IEEE}
}

@article{Schuman2017,
  title={A Survey of Neuromorphic Computing and Neural Networks in Hardware},
  author={Schuman, Catherine D and Potok, Thomas E and Patton, Robert M and Birdwell, J Douglas and Dean, Mark E and Rose, Garrett S and Plank, James S},
  journal={arXiv preprint arXiv:1705.06963},
  year={2017}
}

@article{YuKairong2025,
  title={{TS-SNN}: Temporal Shift Module for Spiking Neural Networks},
  author={Yu, Kairong and Zhang, Tianqing and Xu, Qi and Pan, Gang and Wang, Hongwei},
  journal={arXiv preprint arXiv:2505.04165},
  year={2025}
}

@article{Seekings2025,
  title     = {Integrated Algorithm and Hardware Design for Hybrid Neuromorphic Systems},
  author    = {Seekings, James and Ardakani, Mahsa and Chandarana, Peyton and Eslami, Arshia and Mohammadi, Mohammadreza and Zand, Ramtin},
  journal   = {npj Unconventional Computing},
  volume    = {2},
  number    = {1},
  pages     = {20},
  year      = {2025},
  month     = aug,
  publisher = {Nature Portfolio},
  doi       = {10.1038/s44335-025-00036-2}
}

@ARTICLE{Zeydan2025,
  author={Zeydan, Engin and Alwis, Chamitha De and Khan, Rabia and Turk, Yekta and Aydeger, Abdullah and Gadekallu, Thippa Reddy and Liyanage, Madhusanka},
  journal={IEEE Open Journal of the Communications Society}, 
  title={Quantum Technologies for Beyond 5G and 6G Networks: Applications, Opportunities, and Challenges}, 
  year={2025},
  volume={6},
  number={},
  pages={6383-6420},
  doi={10.1109/OJCOMS.2025.3591842}}

@article{Barbieri2025,
  title={The Security of Quantum Computing in {6G}: From Technical Perspectives to Ethical Implications},
  author={Barbieri, Luca and Menina, Abdelkrim and Bassoli, Riccardo and Fitzek, Frank HP},
  journal={arXiv preprint arXiv:2504.10040},
  year={2025}
}

@article{Narottama2023,
  title={Quantum Machine Learning for {Next-G} Wireless Communications: Fundamentals and the Path Ahead},
  author={Narottama, Bhaskara and Mohamed, Zina and A{\"\i}ssa, Sonia},
  journal={IEEE Open Journal of the Communications Society},
  volume={4},
  pages={2204--2224},
  year={2023},
  publisher={IEEE}
}

@article{Bekolay2014,
  title={Nengo: a Python tool for building large-scale functional brain models},
  author={Bekolay, Trevor and Bergstra, James and Hunsberger, Eric and others},
  journal={Frontiers in Neuroinformatics},
  volume={7},
  pages={48},
  year={2014},
  publisher={Frontiers}
}

@article{Hazan2018,
  title={BindsNET: A machine learning-oriented spiking neural networks library in Python},
  author={Hazan, Hananel and Saunders, Daniel J and Khan, Hassaan and others},
  journal={Frontiers in Neuroinformatics},
  volume={12},
  pages={89},
  year={2018},
  publisher={Frontiers}
}

@article{Stimberg2019,
  title={Brian 2, an intuitive and efficient neural simulator},
  author={Stimberg, Marcel and Brette, Romain and Goodman, Dan FM},
  journal={eLife},
  volume={8},
  pages={e47314},
  year={2019},
  publisher={eLife Sciences Publications, Ltd}
}
\newpage
\end{document}